# A Multi-level Information Integration Framework for Physically Verifiable Fault Diagnosis of Rotating Machinery

## Authors

Yuntong Chen[a], Jianyu Liu[a], Yingqi Li[a], Guobin Zhao[b], Ziang Wang[a], Chao Chen[c], Xitian Tian[a], Lijiang Huang[a*]

a School of Mechanical Engineering, Northwestern Polytechnical University, Xi'an, 710072, China

b Department of Chemistry, National University of Singapore, 3 Science Drive 3, Singapore 117543, Singapore

c School of Mechanical Engineering and Automation, Beihang University, Beijing, 100191, China

E-mail addresses: chenyt123@mail.nwpu.edu.cn(Yuntong Chen), liujy3@mail.nwpu.edu.cn(Jianyu Liu), yingqili@mail.nwpu.edu.cn (Yingqi Li), guobinzhao@nus.edu.sg(Guobin Zhao), wangzi_ang@mail.nwpu.edu.cn(Ziang Wang), chao_chen@buaa.edu.cn(Chao Chen), tianxt@nwpu.edu.cn(Xitian Tian)

**Corresponding author: Prof. Lijiang Huang** (huanglj@nwpu.edu.cn)

## Abstract

Integrating multi-level information, from physical models through data-driven diagnostics to natural language reasoning, into verifiable decision chains is a growing need in intelligent manufacturing. Rotating machinery maintenance, where diagnostic outputs directly inform safety-critical actions, presents a demanding scenario for such integration. In bearing fault diagnosis, taken here as a representative testbed, the standard output is a class label and a confidence score derived from the classifier's own distribution. Extending this output to include quantities comparable against independent physical knowledge would enable per-sample physical verification. Meanwhile, as language models are increasingly adopted for maintenance communication, ensuring that every generated statement remains traceable to verified evidence becomes equally important. This work addresses both aspects from the output side. The proposed Diagnostic Evidence Network (DENet) is an encoder-agnostic multi-task framework that extends the output to a structured evidence record: the classification, a predicted characteristic frequency comparable against the theoretical value determined by bearing geometry and shaft speed, and a temporal localization of transient impulses inspectable on the raw waveform. Across four encoders and three public datasets, this evidence incurs no statistically significant accuracy cost, with a frequency error of about 6 Hz on 1,024-point segments. The deviation between predicted and theoretical frequency constitutes a label-free, inference-time validation signal. It detects misclassifications with AUROC of 0.970 and 0.871, and retains separation within the high-confidence subset. Finally, a QLoRA-adapted language model renders DENet's evidence into traceable maintenance reports without contributing diagnostic decisions, reducing unsupported-claim rates from 10–12% to 2% with no fabricated quantities observed in the evaluated sample.

# 1. Introduction

In modern industrial systems, operational decisions increasingly rely on the integration of heterogeneous information sources at different levels of abstraction. Rotating machinery, which underpins safety-critical operations in aerospace, energy, and manufacturing, presents a particularly demanding scenario for such integration. Among its components, rolling element bearings are particularly failure-prone, and their fault diagnosis from vibration signals has accordingly been studied intensively for decades [1,2]. Driven by deep learning, classification accuracy on widely used benchmarks has reached high levels [3,4]. Industrial adoption, however, has not kept pace, and recent work points to the trustworthiness of model outputs as a contributing factor [5,6]. The standard output is a class label with a softmax confidence score. The confidence score summarizes the classifier's output distribution and is not designed to measure consistency with external physical quantities. As noted in [7], ensuring the reliability and verifiability of diagnostic outputs is valuable for deploying models in safety-critical systems. Meanwhile, large language models increasingly employed to communicate diagnoses may introduce statements that cannot be traced to a verifiable source. This motivates attention to the verifiability of outputs, both the quantities and the language.

Existing research addresses this bottleneck from five directions. The first continues the evolution of encoder architectures, from convolutional networks [8] through Transformers [9] to Mamba-based diagnosers [3]. These methods steadily improve accuracy and efficiency, while the output interface, namely a label and a confidence score, is typically retained. The second applies post-hoc attribution such as Grad-CAM and SHAP [10,11] to trained diagnosers [12], revealing which input regions drive a decision. However, an attribution map does not itself provide a theoretical reference value for comparison, so assessing its physical plausibility typically involves expert visual inspection.

The third direction injects physical knowledge into training, through interpretable kernels [13], wavelet-constrained objectives [14], or physics-guided probabilistic labels [15]. The injected quantities regularize the representation during training, though they are not returned to the user alongside the classification output. The fourth direction accepts the label–confidence interface but seeks to make the confidence itself trustworthy, through Bayesian and evidential uncertainty quantification [5,6]. These methods demonstrably improve calibration, and the resulting estimate is derived from the model's own predictive distribution, to which an external reference is complementary. The fifth employs large language models, as diagnosers operating on measurement data [16] or as question-answering front-ends [17]. These systems improve accessibility, while tracing individual generated statements to a verifiable source remains an open issue. To our knowledge, these directions do not yet deliver a per-sample quantity that is directly comparable against an independently known physical reference. This work explores whether such a comparison can be embedded into the diagnostic output itself.

This work approaches the problem from the output side. We propose the Diagnostic Evidence Network (DENet), a multi-task framework that extends the output from a label–confidence pair to a structured evidence record with three quantities, namely the fault classification, a predicted characteristic frequency comparable against the theoretical value determined by bearing geometry and shaft speed, and a temporal localization of transient impulses inspectable against the raw waveform. The predicted frequency provides complementary information to the confidence score. Its deviation from the theoretical value acts as a per-prediction consistency metric that detects misclassifications with AUROC 0.970/0.871 (vs. 0.983/0.969 for softmax confidence) and retains separation within the high-confidence subset, at no extra inference cost and without ground-truth labels. DENet is encoder-agnostic; a bidirectional multi-scale Mamba encoder (BiMS-Mamba) is adopted as the default backbone for its linear complexity and its fit to the two temporal scales of bearing fault signatures. A QLoRA-adapted language model then generates natural-language maintenance reports grounded exclusively in the evidence record, contributing no diagnostic decisions of its own, so that each statement is designed to map to a specific evidence field.

The main contributions of this work are summarized as follows.

(1) DENet, an encoder-agnostic evidence framework that extends the diagnostic output from a label–confidence pair to per-sample physical evidence, including characteristic frequency prediction and transient impulse localization, without measurable accuracy degradation. The frequency head attains an MAE of about 6 Hz on 1,024-point segments.

(2) A reliability signal that references no classifier statistics. The deviation between predicted and theoretical frequency detects misclassifications with AUROC 0.970/0.871 on two public datasets (one mixed-speed) and provides complementary separation in the high-confidence subset.

(3) An evidence-constrained reasoning scheme, in which a QLoRA-adapted LLM generates reports grounded exclusively in DENet's evidence, with each statement designed to map to a specific evidence field. > Detailed prompt-level prohibitions still leave 10–12% of reports containing unsupported claims, whereas lightweight fine-tuning reduces this to 2%, with no fabricated quantities observed.

The remainder of this paper is organized as follows. Section 2 reviews related work. Section 3 details the BiMS-Mamba encoder, the DENet framework, and the report generation pipeline. Section 4 presents classification benchmarks, ablation and encoder-agnostic studies, diagnostic evidence analysis, and report quality evaluation on three public datasets. Section 5 concludes the paper.

# 2. Related work

This section reviews related work in three groups: encoder architectures (Section 2.1), interpretability and physical knowledge integration (Section 2.2), and uncertainty quantification together with LLM-based diagnosis (Section 2.3), examining each against the three information levels introduced above and a single criterion: whether the quantity delivered to the user can be checked against a reference independent of the model itself. Section 2.4 summarizes the inter-level gaps.

## 2.1 Deep learning methods for machinery fault diagnosis

Deep learning has progressively replaced hand-crafted feature engineering in vibration-based fault diagnosis. Convolutional architectures operating directly on raw signals supported strong anti-noise performance and domain adaptability [8], and attention-based designs further improved feature fusion across time and frequency domains, exemplified by cross-attention Transformers [9]. More recently, selective state space models (SSMs) have entered the field: VibrMamba [3] introduced a lightweight bidirectional Mamba backbone for one-dimensional vibration signals, MD-BiMamba [18] combined modal decomposition with bidirectional feature fusion for aero-engine inter-shaft bearings, and WCamba [19] pursued lightweight deployment for aero-engine diagnosis. These SSM-based diagnosers inherit the linear O(T) complexity of Mamba while matching or exceeding Transformer accuracy, and they represent the current frontier of encoder design.

Two observations emerge. First, classification accuracy on the standard benchmarks has reached high levels [3,4,9], leaving limited room for further improvement on this axis alone. Second, and more importantly for this work, the evolution from CNNs to SSMs has been an evolution within the data-driven inference level: the output interface has remained a class label accompanied by a softmax confidence score, which carries no quantity from the physical-model level against which the prediction can be independently checked.

## 2.2 Interpretability and physical knowledge integration

Existing methods attempt to connect physical knowledge with data-driven diagnostics from two directions. The first applies post-hoc attribution to trained diagnosers to reveal which physical features drive a decision. Grad-CAM [10] and SHAP [11], originally developed for vision and tabular models, have been adapted to vibration diagnosis through multilayer gradient attribution [12], interpretable time-frequency network designs for bearing diagnosis [20], and studies relating SHAP values to fault characteristic frequencies [21]. These methods reveal which input regions or frequency bands drive a decision [22], and in favorable cases the highlighted bands coincide with known fault frequencies. A practical consideration, however, is that an attribution map does not itself provide a theoretical reference value, so assessing its physical plausibility typically involves expert visual inspection.

The second group builds physical knowledge into the model itself. WaveletKernelNet [13,23] parameterizes the first convolutional layer as interpretable wavelet kernels; variational attention has been used to construct intrinsically interpretable Transformers [24]; wavelet-constrained objectives [14], physics-semantics guided domain generalization frameworks [25], and physics-guided probabilistic label structures [15] inject fault mechanism knowledge into training. These designs demonstrably regularize the representation and can improve robustness. The injected quantities, such as kernel center frequencies, envelope constraints, and probabilistic priors, serve as internal regularizers during training and are typically not reported per sample at inference.

A related line employs multi-task learning, attaching auxiliary tasks such as operating-condition recognition or severity identification to a shared encoder [26,27]. These designs indicate that physically meaningful auxiliary objectives can coexist with classification. In these designs, the auxiliary tasks primarily serve the primary task. In terms of information flow, these existing methods leverage physical knowledge within the training process, but the bridge between the physical-model level and the data-driven output level remains one-directional and training-internal: the injected quantities typically do not appear in the per-sample output delivered to the user at inference time. This work instead treats the auxiliary outputs, namely the characteristic frequency and impulse localization, as deliverables retained at inference as per-sample evidence.

## 2.3 Towards trustworthy diagnosis: uncertainty quantification and LLM-based approaches

Within the data-driven inference level, a further line of work accepts the label-plus-confidence interface but seeks to make the confidence itself trustworthy. Bayesian and evidential formulations quantify predictive uncertainty and calibrate confidence [5,6,28–30], and such calibration is widely regarded as a prerequisite for deployment in safety-critical settings [7,31]. An emerging direction has begun to leverage large language models for industrial fault diagnosis under variable operating conditions [32,33]. In the broader machine learning literature, misclassification detection has been formalized around scores derived from the classifier's output distribution, including maximum softmax probability [34] and energy scores [35], as well as selective classification frameworks that trade coverage for risk [36]. These approaches are effective and, in the case of MSP, remarkably strong baselines. These approaches share the property that every score is derived from the classifier's own output distribution, so a confidently incorrect prediction is difficult to flag from these scores alone. The frequency-deviation signal proposed in this work is complementary rather than substitutive. It requires no additional forward pass, compares a model output against an external physical constant, and retains separation within the high-confidence subset (Section 4.5.1).

At the semantic communication level, large language models have been increasingly employed to render diagnostic outputs into natural language. LLMs have been employed as diagnosers that ingest encoded measurement data [16], with systematic fine-tuning studies on complex systems [37], adaptive architectures for variable operating conditions [32], multimodal reasoning frameworks for explainable bearing diagnosis [38,39], and vision-language question-answering front-ends [17,40]; further variants use LLMs for data augmentation [41] or joint diagnosis and prognosis in a single generative model [42]. Recent surveys chart a rapid expansion of this direction [43–45]. These systems markedly improve accessibility for non-expert users. A remaining challenge is faithfulness: when the diagnosis is coupled to a generative process, individual statements in the output may be difficult to attribute to a verifiable source, and unsupported content may be introduced into diagnostic reports. In the adjacent domain of radiology, a systematic review of LLM-generated reports identified similar concerns, with fine-tuned models outperforming prompted ones on faithfulness and expert alignment [46]. Within fault diagnosis, quantitative evaluation of report faithfulness remains limited, and the division of labor between the neural diagnoser and the language model is rarely made explicit.

## 2.4 Summary of research gaps

Within the broader context of industrial information integration [47], the reviewed work has advanced each information level individually. Encoder architectures have steadily improved classification accuracy at the data-driven level (Section 2.1). Attribution and physics-informed methods have demonstrated that physical knowledge can effectively regularize data-driven representations during training (Section 2.2). Uncertainty quantification has improved confidence calibration within the classifier's output distribution, and LLM-based systems have enhanced the accessibility of diagnostic results for non-expert users (Section 2.3). A common characteristic of these advances is that each operates primarily within a single information level. The output delivered to the user typically consists of a class label and a confidence score, both derived from the data-driven level, while physical quantities used during training are not separately reported per sample at inference, and the mapping from diagnostic evidence to generated language is not yet structurally constrained. Whether the diagnostic output can be extended to include a per-sample quantity directly comparable against an independently known physical reference, and whether the mapping from diagnostic evidence to generated language can be structurally constrained, remain open questions. This work explores both directions: embedding a physically comparable quantity into the diagnostic output and grounding report generation in a structured evidence record.

**Table 1. Comparison of representative intelligent fault diagnosis methods with respect to output verifiability.**

| Method | Type | Cls | Freq. evidence | Temporal loc. | Reliability signal | NL report | LLM role |
|---|---|---|---|---|---|---|---|
| WDCNN [8] | CNN | ✓ | ✗ | ✗ | Softmax (internal) | ✗ | — |
| Twins Transformer [9] | Transformer | ✓ | ✗ | ✗ | Softmax (internal) | ✗ | — |
| VibrMamba [3] | Mamba | ✓ | ✗ | ✗ | Softmax (internal) | ✗ | — |
| MD-BiMamba [18] | Mamba | ✓ | ✗ | ✗ | Softmax (internal) | ✗ | — |
| WCamba [19] | Mamba | ✓ | ✗ | ✗ | Softmax (internal) | ✗ | — |
| MLIFN [48] | Multi-source fusion | ✓ | ✗ | ✗ | Softmax (internal) | ✗ | — |
| Multilayer Grad-CAM [12] | Post-hoc XAI | ✓ | ✗ | Saliency map (no reference value) | Softmax (internal) | ✗ | — |
| WaveletKernelNet [13] | Interpretable kernel | ✓ | Kernel parameters (training-internal) | ✗ | Softmax (internal) | ✗ | — |
| PI probabilistic network [15] | Physics-informed | ✓ | Used in training | ✗ | Posterior (internal) | ✗ | — |
| Evidential GRU [28] | UQ | ✓ | ✗ | ✗ | Evidential uncertainty (internal) | ✗ | — |
| FD-LLM [16] | LLM | ✓ | ✗ | ✗ | Token likelihood (internal) | ✓ | Diagnostician |
| FaultGPT [17] | VLM | ✓ | ✗ | ✗ | — | ✓ | Diagnostician |
| FR-LLM [42] | LLM (multi-task) | ✓ | ✗ | ✗ | — | ✓ | Diagnostician |
| Ours | Mamba + LLM | ✓ | ✓ (vs. theoretical value from | ✓ (inspectable on raw | External: physical deviation | ✓ | Evidence-constrained reasoner |

| Method | Type | Cls | Freq. evidence | Temporal loc. | Reliability signal | NL report | LLM role |
|---|---|---|---|---|---|---|---|
| | | | bearing geometry) | waveform) | \|Δf\| | | |

Notes: Reliability signal refers to the quantity available at inference time for judging whether an individual prediction should be trusted. Internal signals are derived from the model's own output distribution, whereas the proposed |Δf| references a theoretical constant determined by bearing geometry and shaft speed, external to the model. Freq. evidence and Temporal loc. indicate whether frequency-level and time-domain evidence are delivered to the user at inference, rather than merely used during training. LLM role distinguishes whether the language model makes diagnostic decisions (Diagnostician) or performs evidence-constrained reasoning that articulates the physical logic from evidence to diagnosis without contributing diagnostic decisions of its own (Evidence-constrained reasoner).

# 3. Methodology

This section presents the proposed multi-level information integration framework. As illustrated in Fig. 1, the framework consists of three modules: (1) the BiMS-Mamba encoder for feature extraction, (2) the Diagnostic Evidence Network (DENet) for physics-referenced multi-task learning, and (3) a domain-adapted LLM for evidence-constrained diagnostic reasoning. The encoder serves as the default backbone and is replaceable (Section 4.4); the evidence-generating components are the contribution of this work.

## 3.1 Overall Framework

As illustrated in Fig. 1, the proposed framework integrates three information levels through three dedicated modules.Given a raw vibration signal $\mathbf{x} \in \mathrm{R}^{1\times T}$ of length $T$, the proposed framework produces three outputs: a fault classification label $\hat{y}_{cls}$, a predicted characteristic frequency $\hat{y}_{freq}$, and a temporal impact probability map $\hat{y}_{imp} \in [0,1]^{T}$. These outputs are further assembled into a structured diagnostic evidence record and rendered into a natural language diagnostic report by a fine-tuned LLM.

The overall pipeline is formulated as:

$$\mathbf{h}_{global},\mathbf{h}_{seq}=f_{enc}(\mathbf{x};\theta_{enc}) \tag{1}$$

$$\hat{y}_{cls},\hat{y}_{freq},\hat{y}_{imp}=f_{DENet}(\mathbf{h}_{global},\mathbf{h}_{seq};\theta_{DENet}) \tag{2}$$

$$\mathbf{r}=f_{LLM}(Evidence(\hat{y}_{cls},\hat{y}_{freq},\hat{y}_{imp});\theta_{LLM}) \tag{3}$$

where $f_{enc}$ denotes the BiMS-Mamba encoder, $f_{DENet}$ denotes the Diagnostic Evidence Network, and $f_{LLM}$ denotes the report generation module. $\mathbf{h}_{global} \in \mathrm{R}^{d}$ is the global feature representation used for classification and frequency regression, $\mathbf{h}_{seq} \in \mathrm{R}^{T\times d}$ is the sequence-level representation used for impact localization, and $\mathbf{r}$ is the generated natural language report. The structured evidence record output by DENet serves as the integration interface between the data-driven and semantic levels, ensuring that the language model receives a fixed-schema input rather than free-form features.

The encoder and DENet are trained jointly in an end-to-end manner with a multi-task loss. The LLM is fine-tuned separately on structured evidence-report pairs generated from the trained model. This two-stage design ensures that all diagnostic decisions are finalized at the data-driven level before the semantic-level module is invoked.

## 3.2 BiMS-Mamba Encoder

The BiMS-Mamba encoder is designed to capture both local transient patterns (e.g., impact impulses) and global temporal dependencies (e.g., periodic fault signatures) in vibration signals. It incorporates three key design choices: selective state space modeling, bidirectional scanning, and multi-scale input processing.

### 3.2.1 Selective State Space Model

The core building block of the encoder is the Mamba block [50], based on the Selective State Space Model (S6). The input is first expanded through a linear projection and a 1-D convolution with SiLU activation, yielding z. Unlike conventional SSMs with fixed parameters, the selective mechanism computes the state-space parameters as functions of the input:.

Given an input sequence $\mathbf{x}\in \mathrm{R}^{T\times D}$, the Mamba block first projects the input to an expanded dimension through a linear layer and a 1-D convolution:

$$\mathbf{z}=\sigma(Conv1d(Linear(\mathbf{x}))) \tag{4}$$

where $\sigma$ denotes the SiLU activation function. The expanded representation is then processed by the selective SSM, which computes input-dependent parameters:

$$\mathbf{B}t=LinearB(\mathbf{z}t),\quad \mathbf{C}t=LinearC(\mathbf{z}t),\quad \Delta_t=softplus(Linear\Delta(\mathbf{z}t)) \tag{5}$$

The continuous-time state space equation is discretized using the zero-order hold (ZOH) rule:

$$\bar{\mathbf{A}}=exp(\Delta_t\mathbf{A}),\quad \bar{\mathbf{B}}t=(exp(\Delta_t\mathbf{A})\text{-}\mathbf{I})(\Delta_t\mathbf{A})^{-1}\Delta_t\mathbf{B}t \tag{6}$$

The hidden state is updated recurrently:

$$\mathbf{h}t=\bar{\mathbf{A}}\mathbf{h}t\text{-}1+\bar{\mathbf{B}}t\mathbf{z}t,\quad \mathbf{y}t=\mathbf{C}t\mathbf{h}_t \tag{7}$$

where $\mathbf{A}\in \mathrm{R}^{N\times N}$ is initialized as a HiPPO matrix and N is the state dimension. This formulation achieves linear computational complexity $O(T)$ with respect to sequence length, compared to the quadratic $O(T^2)$ complexity of Transformer-based attention mechanisms.

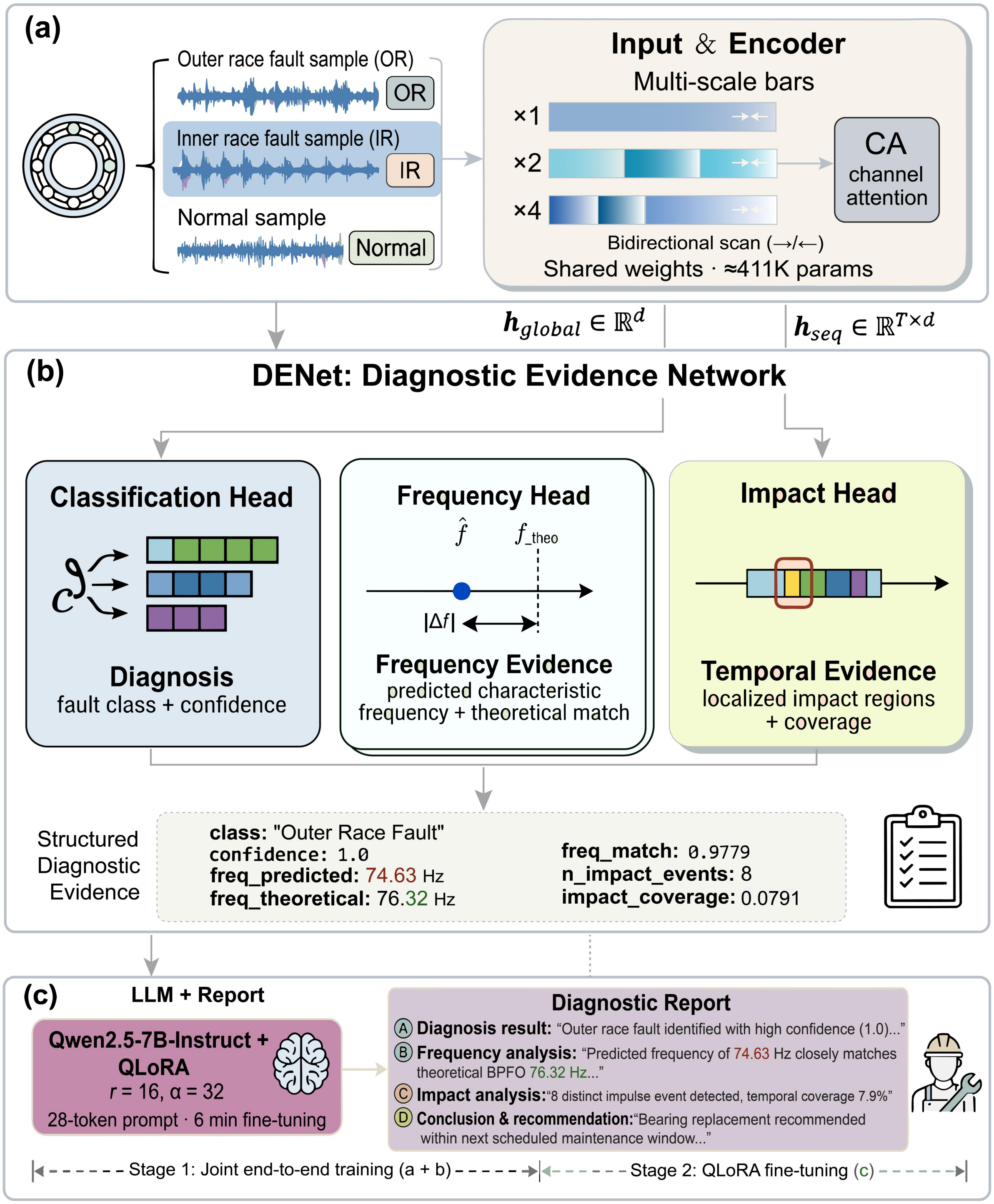

Fig. 1. Overall architecture of the proposed multi-level information integration framework: (a) the BiMS-Mamba encoder, (b) the Diagnostic Evidence Network (DENet) producing the classification (primary task, bold border), the predicted characteristic frequency versus its theoretical value (|Δf|), and the localized transient impulses, assembled into a structured evidence

record, and (c) a QLoRA-adapted LLM that generates a report grounded in the record — all diagnostic decisions are made in (b). Illustrated with the Paderborn three-class task.

3.2.2 Bidirectional Scanning

A single forward scan captures causal dependencies but misses future context that may be informative for fault pattern recognition. To capture both causal and anti-causal context, the BiMS-Mamba encoder employs two independent Mamba branches: a forward branch that scans the signal from $t$=1 to $t$=$T$, and a backward branch that scans the reversed sequence from $t$=$T$ to $t$=1.

Each branch consists of $L$ stacked layers with residual connections and layer normalization:

$$\mathbf{h}fwd^{(l)}=\mathbf{h}fwd^{(l\text{-}1)}+Dropout(MambaBlock(LayerNorm(\mathbf{h}_{fwd}^{(l\text{-}1)}))) \tag{8}$$

$$\mathbf{h}bwd^{(l)}=\mathbf{h}bwd^{(l\text{-}1)}+Dropout(MambaBlock(LayerNorm(\mathbf{h}_{bwd}^{(l\text{-}1)}))) \tag{9}$$

The backward branch output is reversed to align with the original time axis, and the two branches are concatenated:

$$\mathbf{h}bi=[\mathbf{h}fwd^{(L)}|flip(\mathbf{h}_{bwd}^{(L)})]\in \mathrm{R}^{T\times 2D} \tag{10}$$

where | denotes concatenation along the feature dimension and $D$ is the model dimension.

3.2.3 Multi-Scale Input Processing

Bearing fault signatures manifest at different temporal scales: high-frequency impulses occur at the scale of individual sampling points, while the repetition pattern of these impulses reflects the characteristic frequency at a coarser scale. To capture both scales simultaneously, the encoder processes the input signal at multiple resolutions.

Given a set of scale factors $S$=$s_1$,$s_2$,…,$s_K$ (in our implementation, $S$=1,2,4), the input signal is downsampled by each factor:

$$\mathbf{x}^{(k)}=Downsample(\mathbf{x},s_k)\in \mathrm{R}^{1\times\lfloor T/s_k\rfloor} \tag{11}$$

Each downsampled signal is projected to the model dimension via a shared 1-D convolution and processed by the bidirectional Mamba encoder with shared weights:

$$\mathbf{h}^{(k)}=f_{bi\text{-}mamba}(Conv1d(\mathbf{x}^{(k)}))\in \mathrm{R}^{\lfloor T/s_k \rfloor \times 2D} \tag{12}$$

For scales $s_k>1$ , the output is interpolated back to the original length $T$ using linear interpolation to ensure dimensional consistency:

$$\tilde{\mathbf{h}}^{(k)}=Interpolate(\mathbf{h}^{(k)},T)\in \mathrm{R}^{T\times 2D} \tag{13}$$

The multi-scale features are concatenated along the feature dimension and fused via a channel attention mechanism:

$$\mathbf{h}_{cat}=[\tilde{\mathbf{h}}^{(1)}|\tilde{\mathbf{h}}^{(2)}|\cdots|\tilde{\mathbf{h}}^{(K)}]\in \mathrm{R}^{T\times 2KD} \tag{14}$$

$$\mathbf{w}=\sigma(FC2(ReLU(FC1(GAP(\mathbf{h}_{cat}))))) \tag{15}$$

$$\mathbf{h}_{seq}=\mathbf{w}\odot \mathbf{h}_{cat} \tag{16}$$

where GAP denotes global average pooling, $FC1$ and $FC2$ are fully connected layers with a reduction ratio $r$=4, and $\sigma$ is the sigmoid function. The global feature representation is obtained by temporal average pooling:

$$\mathbf{h}_{global}=\frac{1}{T}\sum_{t=1}^{T}\mathbf{h}_{seq,t}\in \mathrm{R}^{2KD} \tag{17}$$

## 3.3 Diagnostic Evidence Network (DENet)

To furnish each diagnosis with checkable evidence, the Diagnostic Evidence Network (DENet) augments the classification head with two physics-referenced auxiliary tasks, namely characteristic frequency regression and impact event localization. Beyond producing auditable evidence, these auxiliary objectives may also influence the shared representation (Section 4.5.3).

3.3.1 Task 1: Fault Classification (Primary Task)

The classification head maps the global feature $\mathbf{h}_{global}$ to class logits through a two-layer MLP:

$$\hat{\mathbf{y}}_{cls}=Linear_2(Dropout(ReLU(Linear_1(\mathbf{h}_{global}))))\in \mathrm{R}^C \tag{18}$$

where $C$ is the number of fault classes. The classification loss is the standard cross-entropy:

$$\mathrm{L}_{cls}=-\sum_{c=1}^{C} y_c \log \hat{p}_c\,, \quad \hat{p}_c=softmax(\hat{\mathbf{y}}_{cls})_c \tag{19}$$

3.3.2 Task 2: Characteristic Frequency Regression (Auxiliary Task)

Each bearing fault type has a theoretically derivable characteristic frequency determined by the bearing geometry and rotational speed. For example, the ball pass frequency of the outer ring (BPFO) is:

$$f_{BPFO}=\frac{n}{2}f_r\left(1-\frac{d}{D}\cos\alpha\right) \tag{20}$$

where $n$ is the number of rolling elements, $f_r$ is the shaft rotation frequency, $d$ is the ball diameter, $D$ is the pitch diameter, and $\alpha$ is the contact angle.

The frequency regression head predicts the normalized characteristic frequency from $\mathbf{h}_{global}$:

$$\hat{y}_{freq}=Linear_2(ReLU(Linear_1(\mathbf{h}_{global})))\in \mathrm{R} \tag{21}$$

A dual-target labeling scheme is adopted: for fault samples, the regression target is the theoretical characteristic frequency of the fault type; for normal samples, the target is the shaft rotation frequency f_r, which is the dominant physically meaningful frequency in a healthy bearing signal. Anchoring normal samples to f_r, rather than excluding them from the regression task, ensures that the frequency head receives a well-defined physical target for every sample and prevents its output on normal samples from drifting arbitrarily. Both predicted and target frequencies are normalized by a dataset-specific maximum frequency constant f_max (200 Hz for both Paderborn and JNU).

Because the normalized targets span roughly an order of magnitude, from the shaft frequency of normal samples (0.05–0.125 after normalization) to the fault characteristic frequencies (0.3–0.62), a plain mean squared error would under-weight the gradient contribution of small targets. The frequency loss is therefore defined as the relative squared error over all samples in the batch:

$$L_{freq}=\frac{1}{B}\sum_i \left(\frac{\hat{y}_{freq}^{(i)}-y_{freq}^{(i)}}{y_{freq}^{(i)}}\right)^2 \tag{22}$$

which places targets of different magnitudes on an equal gradient footing.

To discourage a class-to-frequency lookup, a time-stretch augmentation is applied during training. With probability 0.5, a training segment is temporally rescaled by a factor r drawn uniformly from [0.8, 1.25], and its frequency target is scaled accordingly to r·y_freq. Under this augmentation, samples of the same class carry a continuum of frequency targets, so the label can no longer be predicted from the class identity alone, and the network is encouraged to extract the frequency from the temporal structure of the input. To avoid boundary-padding artifacts, segments are stored with a length of $\lceil 1.25 \times 1{,}024 \rceil = 1{,}280$ points, and the rescaled window is always cropped from real signal before being interpolated to the fixed input length of 1,024 points; per-sample z-score normalization is applied after this final cropping. Validation and test data are never augmented. Two post hoc checks are consistent with this design. On the mixed-speed JNU dataset, the learned head outperforms the best class-wise constant predictor by a factor of ≈1.5 (Section 4.5.1, Table 9). As a sanity check, feeding the frequency head a random vector in place of the encoder feature destroys its prediction accuracy while leaving classification largely intact (Section 4.3, Table 7).

This task serves two purposes. First, the predicted frequency can be compared against the theoretical value to assess physical plausibility. Second, the augmentation encourages frequency-sensitive representations.

### 3.3.3 Task 3: Impact Event Localization (Auxiliary Task)

Bearing faults produce periodic impulse responses in vibration signals. Localizing these impact events provides temporal evidence for the diagnosis. The impact localization head operates on the sequence-level features and outputs a probability of impact occurrence at each time step:

$$\hat{y}_{imp,t}=\sigma(Linear_2(ReLU(Linear_1(\mathbf{h}_{seq,t})))) \in [0,1] \quad (23)$$

The ground truth impulse labels are generated through an automated signal processing pipeline, as illustrated in Fig. 2: (1) Hilbert transform to extract the analytic signal envelope, (2) low-pass smoothing of the envelope using a 4th-order Butterworth filter with a normalized cutoff frequency of 0.1, (3) peak detection on the smoothed envelope with a height threshold of 1.5× the mean value and a minimum inter-peak distance of 20 samples, and (4) binary label assignment within a window of ±5 samples around each detected peak. This pipeline identifies time-domain regions of elevated transient activity, which may arise from fault-induced impulses, structural resonances, or benign mechanical excitation. Importantly, the labels encode the temporal distribution of transient energy rather than fault-specific periodicity; even healthy bearings exhibit transient activity from normal mechanical sources.

Due to the severe class imbalance in impact labels (positive samples typically account for only ~8% of time steps), we employ a dynamically weighted binary cross-entropy loss:

$$\mathrm{L}_{imp}=-\frac{1}{T}\sum_{t=1}^{T}\left[w_t^{+}y_t\log\hat{y}_t+w_t^{-}(1-y_t)\log(1-\hat{y}_t)\right] \quad (24)$$

where the positive class weight $w_t^{+}=(1-\bar{p})/\bar{p}$ and the negative class weight $w_t^{-}=1.0$, with $\bar{p}$ being the mean positive ratio within each batch. This dynamic weighting counteracts the severe class imbalance without requiring manual threshold tuning.

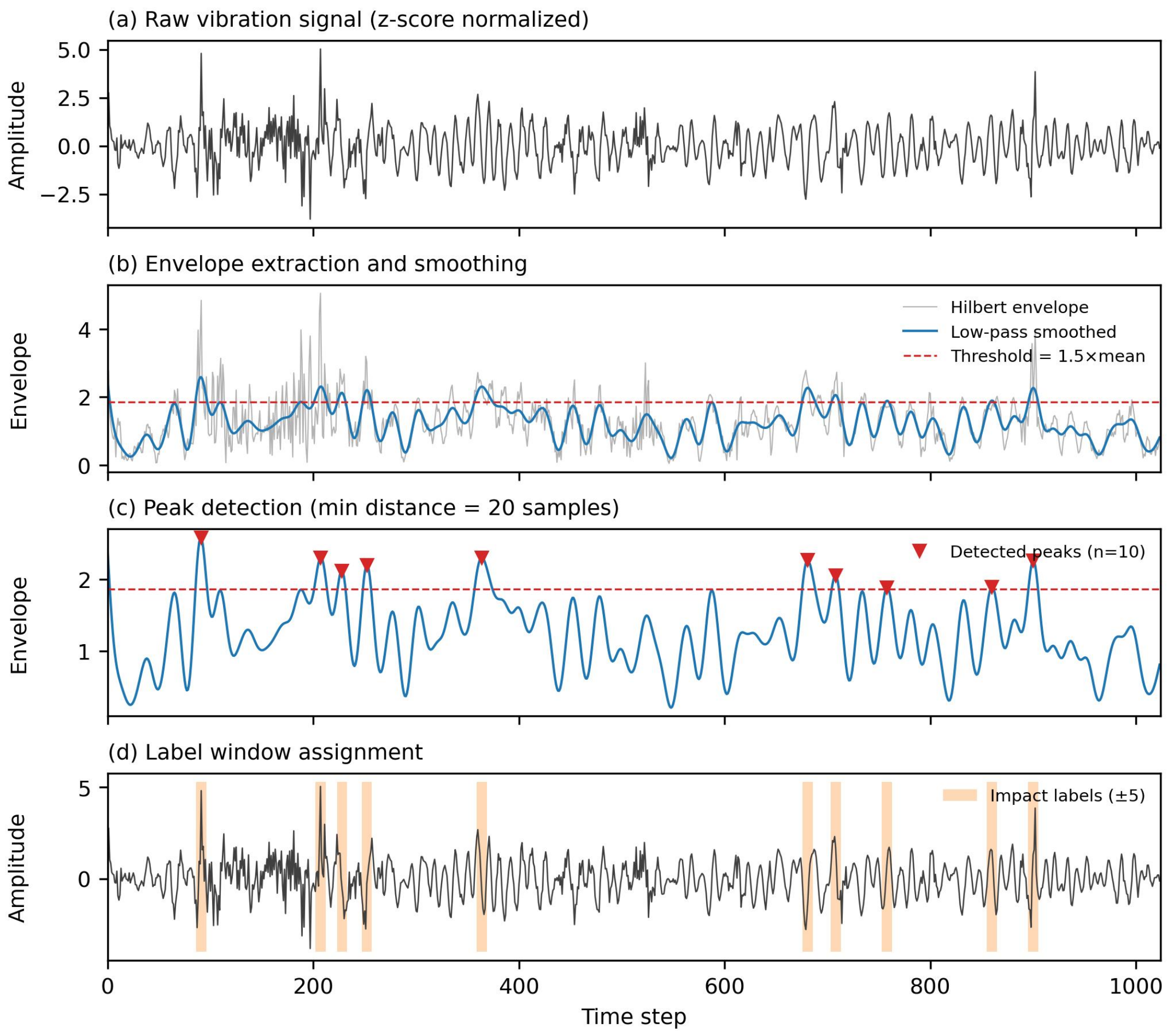

Fig. 2. Automated impulse label generation pipeline illustrated on a representative outer-race fault sample: (a) z-score normalized vibration signal; (b) Hilbert envelope and low-pass smoothed envelope with detection threshold; (c) peak detection; (d) binary label assignment within ±5-sample windows.

### 3.3.4 Multi-Task Training Objective

The overall training loss is a weighted combination of the three task losses:

$$L=L_{cls}+\lambda_{freq}L_{freq}+\lambda_{imp}L_{imp} \tag{25}$$

where $\lambda_{freq}$ and $\lambda_{imp}$ are hyperparameters that control the relative importance of the auxiliary tasks. In our experiments, we set $\lambda_{freq}$=0.05 and $\lambda_{imp}$=0.02, chosen on the Paderborn validation set such that the weighted auxiliary losses remain roughly one order of magnitude below the classification loss throughout training, ensuring the primary task dominates gradient updates.

### 3.3.5 Structured Evidence Generation

After training, the DENet produces a structured diagnostic evidence record for each input signal. This record includes the following fields: (1) the predicted fault class and classifier confidence, (2) the predicted characteristic frequency (in Hz) and its deviation from the theoretical value, enabling direct physical verification, (3) the number of distinct transient impulse regions detected in the signal segment (n_impact_events), the fraction of time steps flagged as impulsive (impact_coverage), and the mean impulse-activity probability across the segment (impact_intensity), which together characterize where and how much transient activity the model has localized. These evidence fields enable a maintenance engineer to audit the model's reasoning, for instance verifying that the predicted frequency matches the expected BPFO for the installed bearing, rather than relying solely on a classification confidence score.

## 3.4 LLM-Based Diagnostic Reasoning and Communication Layer

While the structured evidence from DENet is interpretable to domain experts familiar with vibration analysis terminology, it is less accessible to general maintenance personnel who must make timely decisions based on diagnostic outputs. To bridge this gap, we employ a domain-adapted large language model (LLM) to render the structured evidence into natural language diagnostic reports that are both technically accurate and operationally actionable.

### 3.4.1 Task Formulation

We formulate report generation as a conditional text generation task. Given a structured evidence record e produced by DENet, comprising the predicted fault class, confidence score, predicted characteristic frequency, frequency match score, and impulse localization statistics, the LLM generates a diagnostic report r:

$$r^{*} = argmax\ P(r|\mathbf{e};\theta_{LLM}) \tag{26}$$

The generated report must satisfy three requirements: (1) factual consistency — all numerical values must exactly match the input evidence, with no hallucinated quantities; (2) diagnostic reasoning — the report should articulate the physical logic connecting the evidence to the diagnosis (e.g., explaining that the predicted frequency closely matches the theoretical BPFO, thereby supporting the outer race fault hypothesis); and (3) actionable recommendations — the report should conclude with qualitative maintenance suggestions commensurate with the diagnostic confidence.

It is important to emphasize that the LLM performs evidence-constrained reasoning rather than independent diagnosis. All diagnostic decisions, including classification, frequency prediction, and impact localization, are made by the BiMS-Mamba encoder and DENet. The LLM receives these decisions as structured input and articulates the physical logic connecting the evidence to the diagnosis in readable prose, without contributing diagnostic decisions of its own. This design ensures that diagnostic accuracy is decoupled from the language model's capabilities and that every claim in the generated report is traceable to a specific evidence field.

### 3.4.2 Domain Adaptation via QLoRA

We adopt Qwen2.5-7B-Instruct as the base LLM and fine-tune it using QLoRA (Quantized Low-Rank Adaptation) [51] for parameter-efficient domain adaptation. QLoRA quantizes the pre-trained model weights to 4-bit NF4 precision and trains low-rank adapter matrices injected into the attention layers, reducing GPU memory requirements to approximately 8 GB and enabling fine-tuning on a single consumer-grade GPU.

Training data construction. The fine-tuning dataset consists of 498 evidence-report pairs, constructed through a two-stage pipeline. First, we run the trained BiMS-Mamba + DENet model on the Paderborn dataset to generate structured evidence records for each sample. We then apply stratified sampling to select approximately 500 samples with balanced class distribution (Normal, OR, IR), covering diverse confidence levels and including a small proportion (~10%) of misclassified samples to teach the LLM to express appropriate uncertainty. Second, each structured evidence record is rendered into a reference diagnostic report using a commercial LLM API (Claude Opus 4.6 (Anthropic)) with a carefully designed system prompt that enforces the three requirements defined in Section 3.4.1. The prompt template specifies the report structure (Diagnosis Result → Frequency Analysis → Impact Analysis → Conclusion & Recommendation), mandates the use of exact numerical values from the evidence, and requires professional vibration analysis terminology.

Fine-tuning configuration. LoRA adapters are applied to the query, key, value, and output projection matrices ($\mathbf{W}q$, $\mathbf{W}k$, $\mathbf{W}v$, $\mathbf{W}o$) in all attention layers, with rank $r$=16, scaling factor $\alpha$=32, and dropout rate 0.05. This results in 10.1M trainable parameters (0.13% of the total 7.6B). The model is trained for 3 epochs with a learning rate of $2\times10^{-4}$, cosine learning rate schedule, effective batch size of 16 (micro-batch size 2 with gradient accumulation over 8 steps), and paged AdamW optimizer with 8-bit states. The training converges in approximately 6 minutes on a single NVIDIA RTX 4090 GPU.

# 4. Experiments and results

## 4.1 Experimental Setup

### 4.1.1 Datasets

Three public bearing datasets are employed to validate the proposed framework. Their key specifications and experimental settings are summarized in Table 2.

Table 2. Summary of the datasets and experimental settings.

| Item | CWRU | Paderborn | JNU |
|---|---|---|---|
| Bearing type | SKF 6205-2RS (drive end) | 6203 | N205 / NU205 |
| Sampling rate | 12 kHz | 64 kHz | 50 kHz |
| Operating condition | 0 HP, 1,797 rpm | N15_M07_F10 (1,500 rpm, 0.7 N·m, 1,000 N) | Mixed (600 / 800 / 1,000 rpm) |
| Damage type | Artificial (EDM, single point) | Artificial (EDM, engraving, drilling) | Artificial (wire-cut; 0.3 × 0.25 mm for OR/IR, 0.5 × 0.15 mm for ball) |
| No. of classes | 10 (Normal + IR/OR/Ball × 3 sizes) | 3 (Normal / OR / IR) | 4 (Normal / IR / OR / Ball) |
| Bearings used | — | K001–K006 / KA01,03,05,07,09 / KI01,03,05,07 | N205 (Normal, OR, Ball) / NU205 (IR) |
| Segment length / stride | 1,024 / 512 | 1,024 / 512 | 1,024 / 512 |
| Normalization | Per-sample Z-score | Per-sample Z-score | Per-sample Z-score |
| Total samples | 11,832 (7,099 / 2,366 / 2,367) | 149,972 (59,885 / 50,182 / 39,905) | 17,577 (8,793 / 2,928 / 2,928 / 2,928) |
| Train / Val / Test | 6 : 2 : 2 (stratified) | 89,983 / 29,994 / 29,995 (stratified) | 10,546 / 3,515 / 3,516 (6 : 2 : 2, stratified) |

CWRU dataset. The Case Western Reserve University (CWRU) bearing dataset [4] is collected from drive-end bearings at 12 kHz. Following the common protocol, a 10-class task is constructed under the 0 HP load condition, comprising one healthy condition and nine fault conditions (inner race, outer race, and ball faults, each with three damage diameters of 0.007, 0.014, and 0.021 inches).

Paderborn dataset. The Paderborn University (PU) bearing dataset contains vibration signals from type 6203 ball bearings, with each recording lasting 4 s. A 3-class task is constructed under a single operating condition using the bearings listed in Table 2, all with artificially induced damages produced by electric discharge machining, electric engraving, and drilling. The acceleration signal measured at the bearing housing is used.

JNU dataset. The Jiangnan University (JNU) bearing dataset is collected from cylindrical roller bearings (types N205 and NU205) at a sampling rate of 50 kHz. A 4-class task is constructed comprising one healthy condition and three fault conditions (inner race, outer race, and ball), with artificial damage introduced by wire-cut electrical discharge machining. Data from three shaft speeds (600, 800, and 1,000 rpm) are mixed for training and evaluation, making this the only multi-speed benchmark among the three datasets.

Since CWRU is collected at a single speed and its three damage diameters share the same theoretical characteristic frequency for each fault location, frequency regression degenerates into an approximate class-to-constant lookup; frequency evidence is therefore evaluated only on the Paderborn and JNU datasets.

4.1.2 Implementation Details

The proposed framework is implemented in PyTorch and trained on a single NVIDIA RTX 4090 GPU. The detailed hyperparameter configuration is listed in Table 3. To account for the stochasticity of training, all reported results of BiMS-Mamba are averaged over five independent runs with different random seeds and reported as mean ± standard deviation. The LLM fine-tuning configuration follows Section 3.4.2.

Table 3. Hyperparameter configuration of the proposed framework.

| **Module** | **Parameter** | **Value** |
|---|---|---|
| BiMS-Mamba encoder | Model dimension D | 64 |
| | State dimension N | 16 |
| | Local convolution width | 4 |
| | Expansion factor | 2 |
| | Number of layers L | 4 |
| | Scale factors S | [1, 2, 4] (shared weights) |
| | Dropout rate | 0.1 |
| | Total parameters | ≈ 411k |
| DENet | Loss weight λ_freq | 0.05 |
| | Loss weight λ_imp | 0.02 |
| | Time-stretch range r | [0.8, 1.25] |

| Module | Parameter | Value |
|---|---|---|
| | Augmentation probability | 0.5 |
| | Storage window / output length | 1,280 / 1,024 |
| Impact label generation | Butterworth filter order | 4 |
| | Normalized cutoff frequency | 0.1 |
| | Peak detection threshold | 1.5 × mean(envelope) |
| | Minimum peak distance | 20 samples |
| | Label window half-width w | 5 samples |
| Training | Batch size | 64 |
| | Optimizer / learning rate | Adam / $1\times10^{-3}$ (cosine schedule) |
| | Weight decay | $1\times10^{-4}$ |
| | Max epochs / patience | 100 / 15 |
| | Independent runs | 5 |

4.1.3 Evaluation Metrics

Classification performance is evaluated by accuracy and macro-averaged F1-score. The frequency regression task is evaluated by the mean absolute error (MAE) in Hz between predicted and theoretical characteristic frequencies. Report generation quality is evaluated by the six metrics defined in Section 4.6.1.

## 4.2 Classification Performance Comparison

Although the primary contribution of this work is the construction of an auditable diagnostic evidence chain rather than raw accuracy improvement, it is first necessary to verify that the proposed framework does not sacrifice classification performance. Tables 4–6 report the comparison results on the CWRU, Paderborn, and JNU datasets, respectively.

**CWRU (Table 4).** For fairness and reproducibility, the results of comparison methods are quoted directly from the original publications. BiMS-Mamba achieves 99.81 ± 0.12% accuracy, comparable to the quoted methods including the recent Mamba-based VibrMamba (99.77%).

Nevertheless, given that several recent methods exceed 99.5% on this dataset, the remaining differences lie within statistical noise, and CWRU primarily serves as a sanity check.

**Paderborn (Table 5).** Table 5 includes two quoted multi-source methods (Decision fusion and MLIFN) as reference points, followed by six self-implemented baselines trained under identical settings to ensure a strictly controlled comparison. Among the self-implemented methods, BiMS-Mamba (98.68 ± 0.14%) outperforms the strongest baseline (ResNet, 98.20 ± 0.23%) by 0.48 percentage points and the vanilla Mamba encoder (98.11 ± 0.16%) by 0.57 percentage points, suggesting a modest benefit of the bidirectional multi-scale design over the unidirectional single-scale SSM. The two quoted multi-source methods achieve higher accuracy (98.96% and 99.82%) but exploit complementary torque measurements and four-times-longer input segments; they are included for context rather than direct comparison.

**JNU (Table 6).** The JNU dataset introduces two additional challenges compared with the previous benchmarks: (1) four condition classes, including ball faults, and (2) mixed-speed training across three shaft speeds. The latter broadens the intra-class variability of fault-related frequency components and makes frequency estimation considerably more challenging. All six methods are self-implemented and evaluated under identical settings. Over five runs, BiMS-Mamba achieves a mean accuracy of 99.47 ± 0.11%, slightly above the strongest baseline, ResNet at 99.27 ± 0.07%. The Transformer encoder obtains the lowest baseline accuracy at 98.94 ± 0.20%, followed by Vanilla Mamba at 99.10 ± 0.20%. The representative run used for the confusion-matrix analysis achieves 99.43% accuracy, with 20 misclassified samples out of 3,516 test samples, which is consistent with the five-run average.

As indicated in the last two columns of all three tables, the comparison methods are classification-only. The proposed framework attains comparable accuracy while additionally producing the evidence record, which is examined in detail in Sections 4.4–4.6.

Table 4. Comparison results on the CWRU dataset (10-class classification).

| Method | Accuracy (%) | Freq. | Imp. |
|---|---|---|---|
| 1D-CNN | 99.32 ±0.16 | ✗ | ✗ |
| ResNet | 99.42 ±0.21 | ✗ | ✗ |
| BiLSTM | 98.86 ±0.11 | ✗ | ✗ |

| Method | Accuracy (%) | Freq. | Imp. |
|---|---|---|---|
| Transformer | 99.57 ±0.10 | ✗ | ✗ |
| 1D Mamba | 98.04±0.17 | ✗ | ✗ |
| TRA-ACGAN[3] | 99.70 | ✗ | ✗ |
| MTF-MARN[3] | 99.50 | ✗ | ✗ |
| SE-ResNet[3] | 99.10 | ✗ | ✗ |
| VibrMamba [3] | 99.77 | ✗ | ✗ |
| BiMS-Mamba (ours) | **99.81 ± 0.12** | ✓ | ✓ |

Table 5. Comparison results on the Paderborn dataset.

| Method | Input | Accuracy (%) | Freq. | Imp. |
|---|---|---|---|---|
| 1D-CNN | Vibration | 97.35 ± 0.55 | ✗ | ✗ |
| ResNet | Vibration | 98.20 ± 0.23 | ✗ | ✗ |
| Transformer | Vibration | 97.63 ± 0.42 | ✗ | ✗ |
| BiLSTM | Vibration | 98.09 ± 0.08 | ✗ | ✗ |
| Vanilla Mamba | Vibration | 98.11 ± 0.16 | ✗ | ✗ |
| MLIFN [48] | Vib. + torque | **99.82 ± 0.07** | ✗ | ✗ |
| Decision fusion [49] | Vib. + torque | 98.96 ± 0.29 | ✗ | ✗ |
| BiMS-Mamba (ours) | Vibration | 98.68 ± 0.14 | ✓ | ✓ |

Notes: The two quoted methods (Decision fusion, MLIFN) fuse vibration and torque signals with 4,096-point input segments; the remaining methods use a single vibration channel with 1,024-point segments and are therefore not strictly comparable to the quoted results. All self-implemented baselines share the same training protocol as BiMS-Mamba and differ only in encoder architecture.

Table 6. Comparison results on the JNU dataset (4-class classification, mixed-speed training). All methods are self-implemented under identical settings (5 runs).

| Method | Accuracy (%) | Freq. | Imp. |
|---|---|---|---|
| 1D-CNN | 99.11 ± 0.18 | ✗ | ✗ |
| ResNet | 99.27 ± 0.07 | ✗ | ✗ |
| Transformer | 98.94 ± 0.20 | ✗ | ✗ |
| BiLSTM | 99.22 ± 0.10 | ✗ | ✗ |
| Vanilla Mamba | 99.10 ± 0.20 | ✗ | ✗ |
| BiMS-Mamba (ours) | **99.47 ± 0.11** | ✓ | ✓ |

Confusion matrices. To characterize the per-class behavior of the final model, Fig. 3 presents the confusion matrices obtained from one representative run on the three test sets. The accuracies derived from these matrices therefore correspond to that specific run, whereas Tables 4–6 report the mean and standard deviation over five independent runs.

**On the CWRU dataset**, the representative run misclassifies only 5 out of 2,367 test samples, corresponding to an accuracy of 99.79%. Three of these errors occur among ball-fault classes with different damage diameters: one B021 sample is classified as B007, while two B021 samples are classified as B014. The remaining two errors consist of one IR014 sample classified as IR007 and one B014 sample classified as OR014. Thus, most residual errors arise from distinguishing damage severities within the same fault category rather than from confusion between fundamentally different bearing conditions.

**On the Paderborn dataset**, the representative run achieves an overall accuracy of 98.63%. The dominant confusion occurs between Normal and IR: 193 Normal samples are classified as IR, while 178 IR samples are classified as Normal, forming a nearly symmetric confusion pair. In contrast, OR achieves 99.8% per-class accuracy, with only 19 of 10,037 samples misclassified. This pattern is consistent with the physical observation that mild inner-race damage may produce transient patterns resembling healthy bearing signatures, whereas outer-race faults generally generate more pronounced and spatially localized BPFO-related impulses. The 182 fault samples predicted as Normal (178 IR and 4 OR) constitute missed detections, the costliest error type in maintenance practice, and they fall outside the scope of the frequency-deviation check (see Section 5).

**On the JNU dataset**, the representative run misclassifies 20 out of 3,516 test samples, corresponding to an overall accuracy of 99.43%. The dominant confusion occurs between IR and Ball faults: six Ball samples are classified as IR, while four IR samples are classified as Ball. This is physically plausible because both fault types can produce strong impulsive responses with overlapping spectral components under varying operating conditions. The remaining errors include three IR samples classified as Normal, two OR samples classified as Ball, two Ball samples classified as Normal, and one Ball sample classified as OR. Normal samples are nearly error-free, with only one sample classified as IR and one classified as Ball.

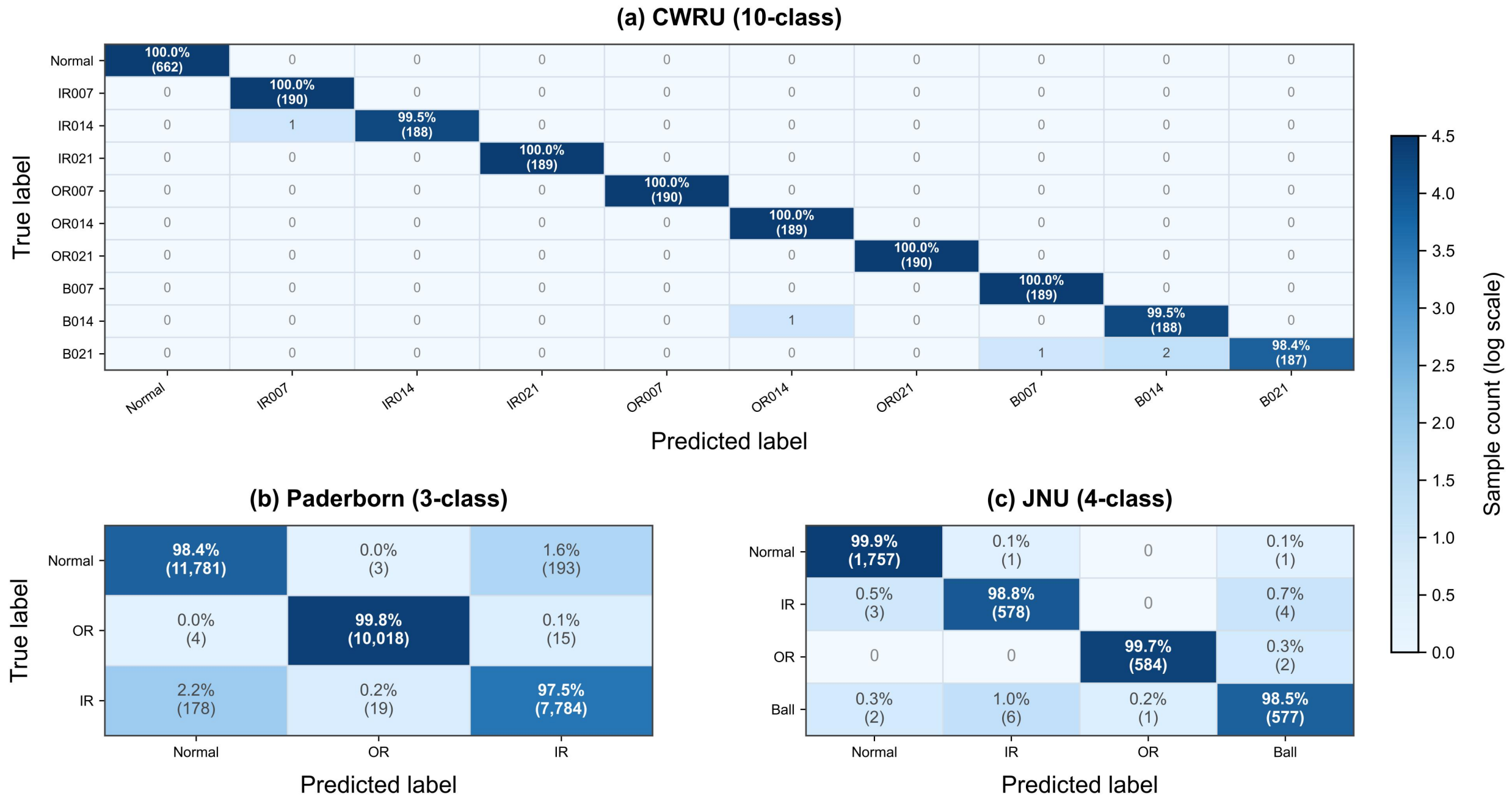

Fig. 3. Confusion matrices on the CWRU (left, 10-class), Paderborn (right top, 3-class), and JNU (right bottom, 4-class) test sets. Diagonal cells report row-normalized classification rates with sample counts; color intensity reflects sample count on a logarithmic scale.

## 4.3 Ablation Study

Both the ablation study and the encoder-agnostic validation (Section 4.4) are conducted on the Paderborn dataset, the largest of the three benchmarks (150k samples), where run-to-run variation is smallest. Cross-dataset results of the full framework are reported in Tables 4–6. To quantify the contribution of each component of DENet, five variants are trained under identical encoder architecture and training configuration, differing only in the auxiliary task losses: (1) the full model; (2) w/o Freq Head (λfreq = 0); (3) w/o Impact Head (λimp = 0); (4) w/o DENet (both auxiliary losses removed, i.e., a conventional classification-only model); and (5) Freq Head w/ Random Input, in which the frequency head receives a random vector instead of the encoder feature, as a sanity check that frequency prediction depends on the encoder feature. Each variant is trained five times with independent random seeds, and results are reported as mean ± standard deviation.

Table 7. Ablation study on the Paderborn dataset (5 runs, mean ± std).

| **Variant** | **Params** | **Accuracy (%)** | **F1 (%)** | **Freq.** | **Imp.** |
|---|---|---|---|---|---|
| Full BiMS-Mamba | 411k | **98.68 ± 0.14** | **98.62 ± 0.15** | ✓ | ✓ |
| w/o Freq Head | 386k | 98.54 ± 0.18 | 98.47 ± 0.19 | ✗ | ✓ |
| w/o Impact Head | 386k | 98.43 ± 0.17 | 98.36 ± 0.18 | ✓ | ✗ |
| w/o DENet (cls only) | 361k | 98.52 ± 0.20 | 98.44 ± 0.21 | ✗ | ✗ |
| Freq Head w/ Random Input | 411k | 98.35 ± 0.19 | 98.27 ± 0.20 | ✗ | ✓ |

Three observations can be drawn from Table 7. First, the full model achieves the highest mean accuracy among all variants, while the random-input variant performs the worst, with a gap of 0.33 percentage points. Given that all differences lie within roughly one standard deviation, we do not claim an accuracy benefit from the auxiliary tasks. The result indicates no measurable accuracy penalty from the auxiliary objectives. Third, the random-input variant still achieves relatively high classification accuracy (98.35%), but loses the ability to predict the characteristic frequency, as expected given that its frequency head receives no signal information. Together with the constant-predictor comparison on the mixed-speed JNU dataset (Table 9), this is consistent with the frequency head using signal information beyond class identity. Overall, all variants lie within a narrow accuracy band of 98.35–98.68%, indicating that the diagnostic evidence capabilities are obtained without statistically significant accuracy degradation, which is consistent with the encoder-agnostic finding in Table 8.

## 4.4 Encoder-Agnostic Validation

The Diagnostic Evidence Network (DENet) is designed as an output framework that is not tied to a specific encoder architecture. To examine this claim, we replace the BiMS-Mamba encoder with three representative architectures — a 1D-CNN (305k params), a Transformer encoder (408k params), and a BiLSTM (612k params) — and train each with and without the DENet auxiliary tasks under identical settings on the Paderborn dataset (5 seeds). The classification-only (cls) variant uses the same encoder with a single classification head, while the +DENet variant adds the frequency and impact heads. The results are reported in Table 8.

Table 8. Encoder-agnostic validation on the Paderborn dataset (5 runs). Δ = (+DENet) − (cls only).

| **Encoder** | **Params (cls / +DENet)** | **cls Accuracy (%)** | **+DENet Accuracy (%)** | **Δ** |
|---|---|---|---|---|
| 1D-CNN | 305k / 338k | 97.35 ± 0.55 | 97.21 ± 0.44 | −0.14 |
| Transformer | 408k / 425k | 97.63 ± 0.42 | 97.46 ± 0.30 | −0.17 |
| BiLSTM | 612k / 645k | 98.09 ± 0.08 | 98.14 ± 0.05 | +0.05 |
| BiMS-Mamba | 361k / 411k | 98.52 ± 0.20 | 98.68 ± 0.14 | +0.16 |

Two conclusions can be drawn from Table 8. First, the absolute accuracy differences |Δ| are only 0.05–0.17 percentage points, which are comparable to or smaller than the run-to-run standard deviations of the corresponding encoders. A paired comparison across all 20 runs (4 encoders × 5 seeds) shows no statistically significant accuracy degradation at the $p = 0.05$ level, indicating no measurable cost on classification performance for the four encoders tested. Second, the parameter overhead introduced by DENet is modest, ranging from 17k for the Transformer to 50k for BiMS-Mamba, corresponding to approximately 4.1%–13.9% of the classification-only encoder parameters.

These results support the encoder-agnostic design of DENet. The diagnostic evidence outputs, namely characteristic frequency prediction and impulse localization, can be attached to different sequence encoders without a measurable loss on the primary classification task. The choice of BiMS-Mamba as the default encoder reflects its accuracy, which is the highest among the tested encoders both in the classification-only setting (98.52%) and with DENet (98.68%), as well as its structural suitability for vibration signals, rather than a coupling between DENet and a specific encoder.

## 4.5 Diagnostic Evidence Analysis

This section examines the quality of the structured evidence produced by DENet, which constitutes the core interpretability claim of this work.

### 4.5.1 Characteristic Frequency Prediction

Fig. 4 presents the predicted characteristic frequencies for all fault samples in the Paderborn test set, grouped by classification outcome, with theoretical values indicated by dashed lines (BPFO = 76.32 Hz for outer race faults, BPFI = 123.68 Hz for inner race faults). Over all fault-class samples, the frequency MAE is 1.26 Hz for OR (n = 10,037) and 12.35 Hz for IR (n = 7,981), for an overall MAE of 6.17 Hz. As Fig. 4 shows, predictions for correctly classified samples cluster tightly around the theoretical values, whereas misclassified samples exhibit drastic frequency deviations. The misclassified IR samples in Fig. 4(b) concentrate near the shaft-frequency anchor rather than scattering randomly. Most of them are predicted as Normal, so the classifier and the frequency head err coherently. A consistency check between the two heads exposes errors on which the heads disagree and is blind to errors on which they agree, which delimits the scope of the reliability signal analyzed below. This agreement provides frequency-domain evidence for each diagnosis. A maintenance engineer can directly check whether the predicted frequency matches the expected characteristic frequency of the installed bearing, as a cross-check complementary to the confidence score.

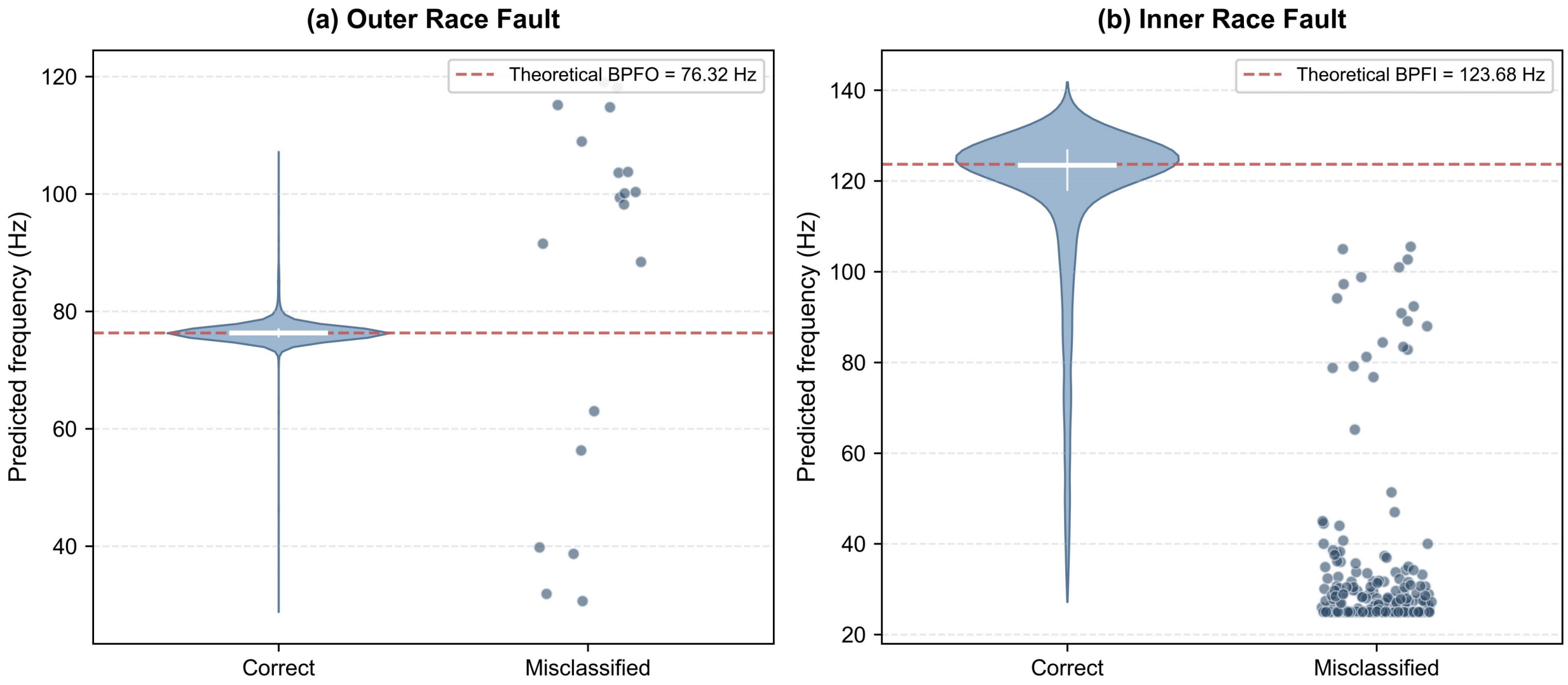

Fig. 4. Predicted characteristic frequencies on the Paderborn test set, grouped by classification outcome: (a) outer race fault, (b) inner race fault. Violin width encodes sample density for correctly classified samples; misclassified samples are shown individually as jittered points. Dashed lines indicate theoretical values (BPFO = 76.32 Hz, BPFI = 123.68 Hz). The thin lower tail of correctly classified IR samples corresponds to segments within the trough of shaft-frequency amplitude modulation (cf. Table 9); misclassified IR samples concentrate near the shaft-frequency anchor of the predicted class. Results are from a representative run (seed 1).

JNU frequency prediction. Table 9 reports the per-class frequency MAE on the JNU test set, which constitutes the most challenging frequency regression benchmark among the three datasets due to mixed-speed training (600 / 800 / 1,000 rpm). Despite the substantially broadened intra-class frequency variance, DENet achieves an overall MAE of 6.10 Hz — comparable to the single-speed Paderborn result — with per-class values of 10.21 Hz (IR), 4.62 Hz (OR), and 3.49 Hz (Ball). The Ball fault class attains the lowest error, while the IR class exhibits the highest error, consistent with the physical observation that inner race fault impulses are amplitude-modulated by the shaft rotation and are therefore more difficult to characterize spectrally in short signal segments. The gap between the mean and median errors in Table 9 reveals that the error structures of the two datasets differ in an instructive way. On Paderborn, the IR error is dominated by a minority of segments: half of the IR predictions deviate by less than 4.21 Hz, and the elevated MAE stems from segments falling within the trough of the shaft-frequency amplitude modulation, where little characteristic rhythm is present in the 16 ms window. On JNU, by contrast, mixed-speed training broadens the IR error more uniformly (median absolute error 8.63 Hz), reflecting the compounded difficulty of inferring both the fault signature and the operating speed from a single short segment.

Table 9. Characteristic frequency prediction on the Paderborn and JNU test sets (fault classes only; MAE / median absolute error).

| **Dataset** | **Condition** | **IR MAE / Med. (Hz)** | **OR MAE / Med. (Hz)** | **Ball MAE / Med. (Hz)** | **Overall MAE / Med. (Hz)** | **Const. baseline MAE (Hz)** |
|---|---|---|---|---|---|---|
| Paderborn | Single speed (1,500 rpm) | 12.35 / 4.21 | 1.26 / 0.78 | — | 6.17 / 1.37 | — |
| JNU | Mixed speed (600/800/1,000 | 10.21 / 8.63 | 4.62 / 3.88 | 3.49 / 2.73 | 6.10 / 4.33 | 9.37 |

| Dataset | Condition | IR MAE / Med. (Hz) | OR MAE / Med. (Hz) | Ball MAE / Med. (Hz) | Overall MAE / Med. (Hz) | Const. baseline MAE (Hz) |
|---|---|---|---|---|---|---|
| | rpm) | | | | | |

Notes: Median absolute error (Med.) is reported alongside MAE because the error distributions are heavy-tailed: the mean is dominated by a minority of large-error segments (see text).

Under mixed-speed training, a label-to-frequency lookup is limited by the spread of per-speed theoretical frequencies. The best class-wise constant predictor attains an MAE of 9.37 Hz, roughly 1.5 times that of DENet (6.10 Hz), indicating that the frequency head uses speed-dependent information beyond a class-to-frequency mapping.

Comparison with spectral resolution limits. To contextualize the above MAE values, Table 10 compares the DENet frequency prediction accuracy against the fundamental frequency resolution of conventional FFT-based spectral analysis applied to the same 1,024-point input segments. The FFT frequency resolution is determined by $\Delta f = f_s / N$, where $f_s$ is the sampling rate and N is the segment length.

Table 10. DENet frequency MAE in the sub-period-segment regime, with the FFT bin width of the same segment length as a scale reference.

| Dataset | f_s (kHz) | Segment duration (ms) | Periods per segment (BPFO) | FFT Δf (Hz/bin) | DENet MAE (Hz) |
|---|---|---|---|---|---|
| Paderborn | 64 | 16.0 | 1.22 | 62.50 | 6.17 |
| JNU | 50 | 20.5 | ≈0.8–1.4 | 48.83 | 6.10 |

Notes: FFT $\Delta f = f_s / N$ is reported as an intuitive scale reference; MAE and bin width are not directly commensurable quantities. Periods per segment = segment duration / theoretical BPFO period (13.10 ms for Paderborn; 14.7–24.4 ms for JNU, spanning the three shaft speeds).

On the Paderborn dataset, each 1,024-point segment spans only 16 ms — approximately 1.2 repetition periods of the BPFO impulse train and fewer than two periods of the BPFI. At this segment length, the 62.5 Hz bin width places the two fault characteristic frequencies (BPFO = 76.32 Hz, BPFI = 123.68 Hz) in adjacent bins, and envelope-spectrum methods typically require multiple impulse repetitions that a 16 ms window may not contain. DENet's output nevertheless lies within 6.17 Hz of the theoretical value on average. Since a 16 ms window contains too few repetitions for periodicity-based estimation, the head presumably relies on class-related and speed-related signal morphology, such as resonance decay rates and modulation patterns. On the JNU mixed-speed dataset the same sub-period regime holds, and a comparable MAE of 6.10 Hz is obtained with three shaft speeds mixed.

This capability is complementary to, not a replacement for, classical envelope spectrum analysis, which achieves sub-hertz accuracy on sufficiently long recordings (e.g., 4-second segments at 64 kHz yield $\Delta f = 0.25$ Hz). Rather, its value lies in providing an embedded physical consistency check that is produced simultaneously with the classification decision on 1,024-point segments, a signal length at which longer-window spectral methods would typically be applied separately. This co-produced frequency estimate provides a cross-check of the classification output without requiring a separate signal processing pipeline. This accuracy level is also commensurate with its intended role as a consistency check. The overall MAE remains below 15% of the spacing between BPFO and BPFI on Paderborn (≈47 Hz), which suffices to separate these two fault hypotheses at the sample level.

Frequency deviation as a complementary reliability signal. Beyond serving as physically verifiable evidence, the predicted characteristic frequency enables an operational reliability check. We define the frequency deviation $|\Delta f|$ as the absolute difference between the predicted frequency and the theoretical characteristic frequency of the predicted class. Since it references the predicted rather than the true label, $|\Delta f|$ is computable at inference time without ground truth, given the bearing geometry and shaft speed.

Table 11(a) benchmarks $|\Delta f|$ against two standard confidence-based misclassification detectors, maximum softmax probability (MSP) and the energy score. Although $|\Delta f|$ makes no reference to the classifier's output distribution, it detects misclassifications with AUROC 0.970/0.871 on Paderborn/JNU (vs. 0.983/0.969 for MSP). Erroneous diagnoses often produce frequency estimates that deviate substantially from the theoretical value of the predicted class, which makes many of them visible. Combining the two signals changes AUROC marginally (0.984/0.971) and lowers FPR@95%TPR on JNU (0.086 vs. 0.149 for MSP alone) but not on Paderborn (0.069 vs. 0.061). Most errors are visible to both signals simultaneously, so the headroom for combination is small. The value of \|Δf\| is as a complementary check, as quantified below.

Table 11(b) makes the stratification explicit. On Paderborn, 79% of fault-class predictions fall within 5 Hz of the theoretical value, where the error rate is only 0.03%, supporting automated acceptance of frequency-consistent diagnoses. The error rate climbs monotonically with $|\Delta f|$ (0.03% → 0.52% → 1.67% → 15.31%), with errors concentrating overwhelmingly in the tail: 205 of the 230 misclassifications fall in the >20 Hz bin. The intermediate bins are populated mainly by correctly classified IR segments belonging to the heavy tail characterized in Table 9, which dilutes but does not break the gradient; the corresponding stratification on JNU spans 0.10% → 10.77%. The stratification also appears within the high-confidence subset. Among samples with softmax confidence exceeding 0.99, diagnoses with $|\Delta f| > 10$ Hz are misclassified roughly 18× (Paderborn, 7/905 vs. 7/15,960) and 10× (JNU, 2/278 vs. 1/1,382) more often than frequency-consistent ones, although the counts are small. These 7 and 2 errors would be accepted by a 0.99 confidence threshold and are flagged by the frequency check.

We emphasize that no causal claim is made: a large $|\Delta f|$ may reflect an intrinsically ambiguous signal degrading both heads simultaneously. Regardless of mechanism, $|\Delta f|$ is a by-product of the diagnosis that stratifies samples by error rate, suggesting a simple escalation rule (e.g., $|\Delta f| > 10$ Hz → expert review) even when the classifier itself reports high confidence.

Table 11. Misclassification detection via frequency deviation on the Paderborn and JNU test sets.

(a) Detection performance of unreliability scores

| Dataset | Score | AUROC | AUPR | FPR@95%TPR |
|---|---|---|---|---|
| Paderborn | MSP | 0.983 | 0.38 | 0.061 |
| | Energy | 0.971 | 0.27 | 0.089 |
| | FreqDev | 0.970 | 0.33 | 0.121 |
| | MSP + FreqDev | 0.984 | 0.39 | 0.069 |
| JNU | MSP | 0.969 | 0.23 | 0.149 |
| | Energy | 0.968 | 0.22 | 0.173 |
| | FreqDev | 0.871 | 0.33 | 0.426 |
| | MSP + FreqDev | 0.971 | 0.32 | 0.086 |

(b) Error rate stratified by $|\Delta f|$

| Dataset | \|Δf\| bin (Hz) | n | Errors | Error rate (%) | High-conf. (n / Errors) | High-conf. rate (%) |
|---|---|---|---|---|---|---|
| Paderborn | 0–5 | 14,315 | 5 | 0.03 | 15,960 / 7 | 0.04 |
| | 5–10 | 1,720 | 9 | 0.52 | | |
| | 10–20 | 658 | 11 | 1.67 | 905 / 7 | 0.77 |
| | >20 | 1,339 | 205 | 15.31 | | |
| | Total | 18,032 | 230 | 1.28 | | |
| JNU | 0–5 | 1,008 | 1 | 0.10 | 1,382 / 1 | 0.07 |
| | 5–10 | 418 | 3 | 0.72 | | |
| | 10–20 | 263 | 4 | 1.52 | 278 / 2 | 0.72 |
| | >20 | 65 | 7 | 10.77 | | |
| | Total | 1,754 | 15 | 0.86 | | |

Notes: FreqDev |Δf| = |predicted frequency − theoretical frequency of the predicted class|, computable at inference time without ground-truth labels. MSP = 1 − maximum softmax probability; Energy = −log Σ exp(logits). The combined score is a logistic regression over rank-normalized MSP and FreqDev, fitted on the validation set only. The analysis covers fault-class predictions (18,032 / 29,995 samples for Paderborn; 1,754 / 3,516 for JNU); Normal predictions are excluded as their regression anchor is the shaft frequency (Section 3.3.2). For JNU, theoretical frequencies follow the measured shaft speed. The high-confidence subset (confidence > 0.99) is stratified into two bins owing to limited errors; with only 15 JNU misclassifications, intermediate-bin rates fluctuate, and the tail concentration (10.77% vs. 0.10%) is the robust finding. Bearing geometry: 6203 (Paderborn), N205/NU205 (JNU, Section 4.1.1).

4.5.2 Transient Impulse Localization

Fig. 5 visualizes the impulse localization results for representative OR, IR, and Normal samples. Time-domain regions where the predicted impulse probability exceeds 0.7 are highlighted on the raw vibration waveform. The highlighted regions align with visually identifiable transient bursts, indicating that the model has learned to localize time-domain intervals of elevated transient energy. The annotated statistics (number of discrete impulse events and temporal coverage) are consistent with the temporal distribution of transient activity encoded by the label generation pipeline (Fig. 2).

It should be noted that the impulse localization provides temporal evidence, showing where transient activity occurs, rather than fault-specific periodicity information. As shown in the Normal panel of Fig. 5, healthy bearings also exhibit measurable impulse activity arising from benign mechanical excitation such as structural resonances. The diagnostic discrimination between fault and healthy conditions rests on the classification head and the frequency regression head, not on the impulse statistics alone. The value of impulse localization lies in pointing the engineer to signal regions of elevated transient energy for inspection.

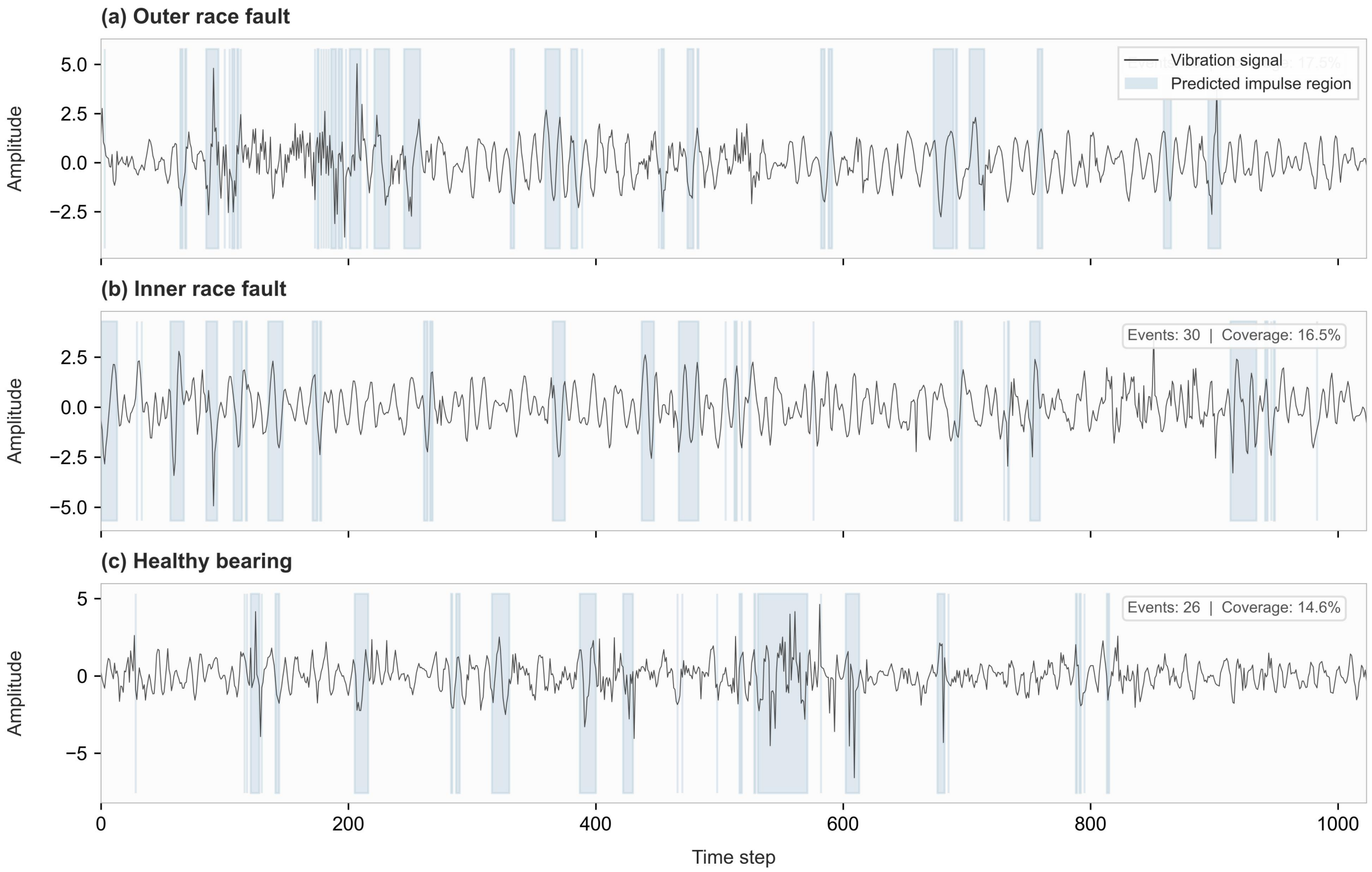

Fig. 5. Impulse localization results for representative samples: (a) outer race fault, (b) inner race fault, (c) healthy bearing. Shaded regions indicate model-predicted impulse events overlaid on the raw vibration waveform.

### 4.5.3 Feature Space Visualization

Fig. 6 compares the t-SNE projections of the learned global features between the classification-only model (w/o DENet) and the proposed full model. Without the physics-referenced auxiliary tasks, the feature clusters exhibit noticeable inter-class overlap, particularly between Normal and IR samples, and each class fragments into loosely connected sub-clusters with diffuse boundaries.

In contrast, the full model produces visibly tighter and better-separated clusters. Sub-clusters within each class possibly reflect differences in damage generation methods (EDM, engraving, drilling) and manufacturing tolerances across test bearings, while inter-class boundaries remain clearly separated. This qualitative observation is consistent with the ablation results.

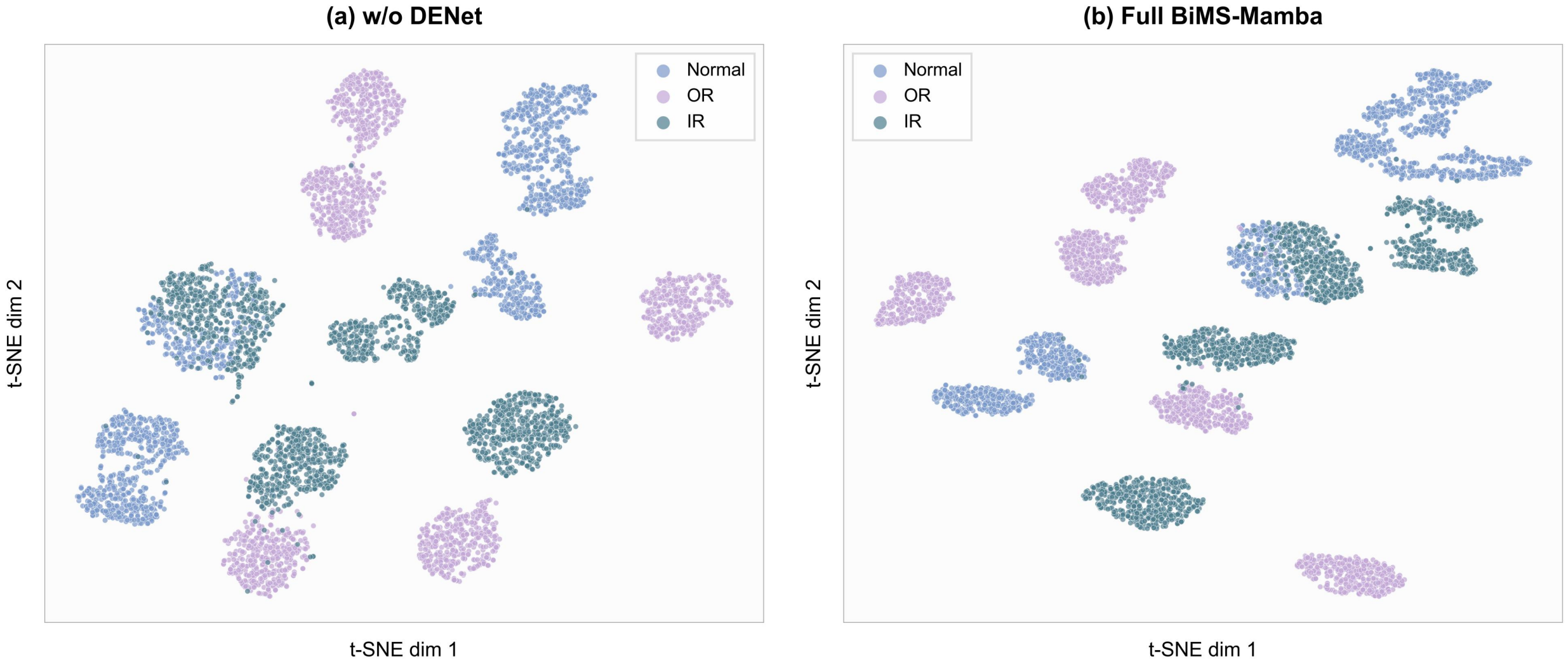

Fig. 6. t-SNE visualization of learned global features on the Paderborn test set. (a) Classification-only model (w/o DENet); (b) Full BiMS-Mamba with DENet. Each point represents a 1,024-point test segment; colors indicate fault classes.

## 4.6 Evidence-Grounded LLM Evaluation

### 4.6.1 Quantitative Comparison

To assess the value of domain-specific fine-tuning, we compare the QLoRA-adapted model against three baselines built on the same base model (Qwen2.5-7B-Instruct), ordered by increasing supervision: (1) ZS-short — the same minimal 28-token system prompt used by the fine-tuned model, specifying the task but no formatting rules or content constraints; (2) ZS-full — the complete eight-rule system prompt used to generate the training data (≈1,100 tokens), the strongest prompt-engineering baseline; and (3) Few-shot — ZS-full plus one training-set exemplar report (≈1,500 tokens). All settings share the same held-out evaluation set — the last 40% of the evidence–report pairs (n = 200), never seen during training — and generation parameters (temperature 0.7, max 600 tokens, one generation per sample).

Reports are assessed by six metrics in two groups (Table 12): a faithfulness group measuring whether generated content stays within the bounds of the input evidence, and an adaptation gain group measuring what fine-tuning adds beyond instruction following; metric definitions and annotation protocol are given in the table notes. The three annotation-based metrics are rated blind to model identity, with the two faithfulness counts independently verified by a second rater.

Table 12. Report generation quality across four settings. ZS = zero-shot; FT = QLoRA fine-tuned. All rates computed over n = 200 held-out samples; report-level counting (a report with multiple violations counts once).

| **Group** | **Metric** | **ZS-short** | **ZS-full** | **Few-shot** | **FT** |
|---|---|---|---|---|---|
| Faithfulness | Format compliance (%) | 0.0 | 100.0 | 100.0 | **100.0** |
| | Numerical consistency (%) | 100.0 | 100.0 | 100.0 | **100.0** |
| | Unsupported claim rate (%) | 18 | 12 | 10 | **2** |
| | Fabricated quantity rate (%) | 10 | 6 | 4 | **0** |
| Adaptation gain | Recommendation calibration (%) | 26 | 54 | 76 | **96** |
| | Style internalization (ROUGE-L, %) | 16.0 | 30.6 | 41.5 | **61.4** |

The results in Table 12 reveal a monotonic supervision ladder across the four settings, with the two metric groups telling complementary stories.

Within the faithfulness group, format compliance and numerical consistency are within the high-confidence subset once detailed instructions are provided: ZS-short produces no structurally compliant reports, indicating that the report format is entirely learned behavior, whereas all other settings reach 100% on both metrics — numerical grounding, in particular, is guaranteed by the structured evidence input rather than by any adaptation. The two stricter faithfulness metrics, however, show a residual gap that instructions did not close in our setting. The unsupported claim rate decreases monotonically with supervision strength (18% → 12% → 10%) yet remains at 10% (20/200) even for the strongest prompt-based setting, despite the system prompt explicitly prohibiting such content. Fine-tuning reduces it to 2%, and no fabricated quantities are observed (0/200 vs. 4–10% for prompt-based settings). This contrast suggests that, for the 7B model tested, behavioral constraints are partially enforced by prompt context but more reliably acquired through fine-tuning. This matters in practice because the residual unsupported claims in prompt-based reports include fabricated remaining-useful-life estimates.

The adaptation-gain group shows the same ladder at larger amplitude. Recommendation calibration improves from 26% to 54% with detailed instructions and to 76% with one exemplar, but even the few-shot setting leaves nearly a quarter of reports recommending actions misaligned with the evidence strength — most commonly, hedged verification requests for diagnoses with confidence above 0.99 and frequency match above 0.95. Fine-tuning raises this to 96% (192/200), suggesting that calibrated phrasing is more readily acquired from examples than from rules. ROUGE-L exhibits the same trend (16.0 → 30.6 → 41.5 → 61.4) and is reported as a supplementary indicator of stylistic internalization rather than factual quality.

From a deployment perspective, the fine-tuned model requires only a 28-token system prompt compared to ≈1,100 tokens for ZS-full and ≈1,500 tokens for Few-shot — a 97% reduction in per-inference prompt overhead. Combined with the six-minute adaptation cost on a single consumer-grade GPU, and considering that no prompt-based setting matches the fine-tuned model on either strict-faithfulness or calibration metrics, QLoRA adaptation is low in cost and, for this 7B base model, more reliable than the prompt-based settings.

4.6.2 Case Study

Fig. 7 presents an end-to-end case study of the complete pipeline for an outer race fault sample, illustrating how the same structured evidence record is rendered, faithfully or otherwise, under three supervision settings.

The DENet evidence record (Fig. 7b) reports an OR classification with confidence 1.0, a predicted characteristic frequency of 74.63 Hz against the theoretical BPFO of 76.32 Hz (match score 0.9779, absolute deviation 1.69 Hz), and one localized impulse event with an impact coverage of 0.0791. By the reliability criteria supported in Section 4.5.1, this constitutes a strong-evidence diagnosis: the classifier is maximally confident, and the frequency deviation falls well within the 5 Hz band where the empirical error rate is lowest.

The three generated reports (Fig. 7c) exhibit the failure hierarchy quantified in Table 12. The ZS-short report, generated without task-specific guidance, produces a free-form narrative that omits the required four-section structure entirely — a structural failure consistent with its 0% format compliance. The ZS-full report follows the prescribed structure and references the numerical values correctly, but contains unsupported claims: it asserts that the frequency evidence is "consistent with resonance behavior" and attributes the localized impulses to a "defect-induced impact source" — physical interpretations that appear in no field of the evidence record and, in the latter case, overstep the interpretive boundary of the impulse labels supported in Section 4.5.2. This content appears despite the ≈1,100-token system prompt explicitly prohibiting it, exemplifying at the level of a single report the residual unsupported-claim rate (10–12%) that instructions alone did not eliminate. Its concluding recommendation also hedges toward further verification despite the maximal evidence strength — the calibration failure mode that dominates the prompt-based settings in Table 12.

The fine-tuned report satisfies all three requirements defined in Section 3.4.1. It follows the four-section structure and references every numerical value faithfully (factual consistency). It articulates the reasoning chain connecting evidence to diagnosis, noting for example that the predicted frequency of 74.63 Hz matches the theoretical BPFO of 76.32 Hz within 1.69 Hz, supporting the outer race fault hypothesis, and describes the localized impulse activity as temporal evidence of transient energy without overclaiming fault-specific periodicity, consistent with the interpretive boundary supported in Section 4.5.2 (diagnostic reasoning). Finally, it concludes with maintenance recommendations commensurate with the strong evidence, recommending scheduled corrective action rather than hedged re-verification, an instance of the calibration behavior measured in Table 12 (actionable recommendations). In this example, every claim in the report is traceable to a specific field of the evidence record, and no content exceeds what the evidence supports, consistent with the evidence-constrained reasoning role defined in Section 3.4.1.

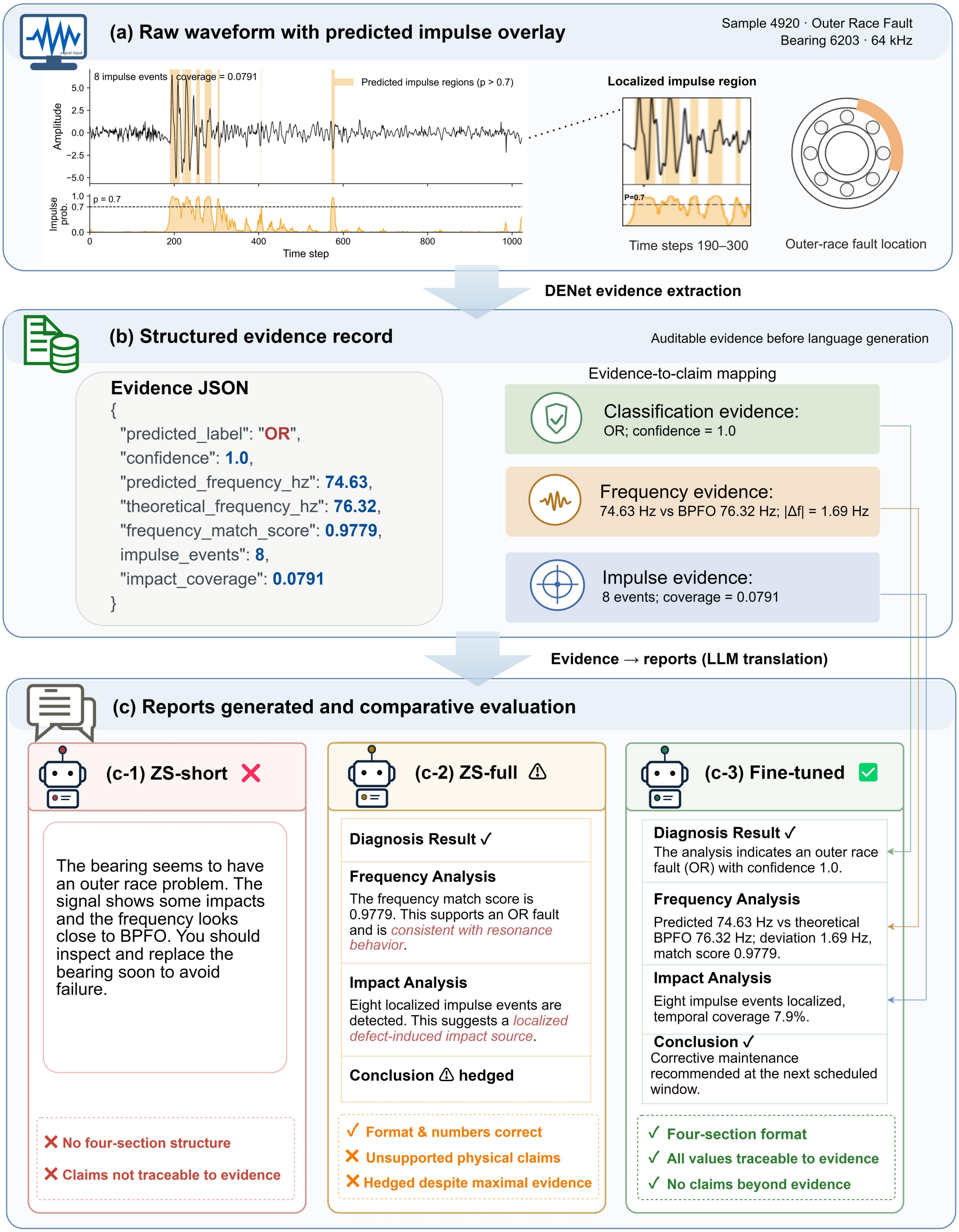

Fig. 7. End-to-end case study for an outer race fault sample (Paderborn, Sample 4920): (a) raw vibration waveform with predicted impulse regions (p > 0.7) overlaid; (b) the structured evidence

record produced by DENet and its mapping to report claims; (c) reports generated from the same evidence record under three supervision settings, ZS-short, ZS-full, and QLoRA fine-tuned, with per-report compliance summarized below each panel. Statements unsupported by the evidence record are marked in red italics.

#### 4.6.3 Expert Evaluation

While the metrics in Table 12 assess format, faithfulness, and lexical fidelity, they do not measure whether the generated reports are useful to practitioners. To this end, a blind expert evaluation is conducted. Three evaluators with backgrounds in vibration analysis and rotating machinery maintenance independently rate 100 report pairs randomly sampled from the 200 held-out samples. For each evidence record, the zero-shot (ZS-full) and fine-tuned reports are presented side by side in randomized order without model identity, and each report is rated on a five-point Likert scale along three dimensions: (1) technical accuracy, whether the stated reasoning is consistent with the evidence and with vibration analysis practice; (2) readability, whether the report is clear and well-organized for maintenance personnel; and (3) actionability, whether the maintenance recommendations are specific and commensurate with the assessed severity. Table 13 summarizes the results.

Table 13. Expert evaluation results (mean ± std, 5-point Likert scale).

| Dimension | ZS-full | Fine-tuned |
|---|---|---|
| Technical accuracy | 3.54 ± 0.16 | 4.50 ± 0.07 |
| Readability | 3.64 ± 0.16 | 4.51 ± 0.04 |
| Actionability | 3.83 ± 0.07 | 4.48 ± 0.04 |

As shown in Table 13, the fine-tuned model outperforms the zero-shot baseline across all three dimensions. The largest gap appears in technical accuracy (4.50 vs. 3.54), where evaluators noted that the fine-tuned reports consistently employ precise vibration analysis terminology and provide physically grounded interpretations — for example, correctly attributing the predicted frequency of Normal samples to non-fault spectral components rather than leaving them unexplained. The actionability gap (4.48 vs. 3.83) reflects a recurring pattern in ZS-full reports: recommending additional verification even when the evidence strongly supports the diagnosis (confidence = 1.0, frequency match > 0.97), whereas the fine-tuned model calibrates its recommendations to the diagnostic certainty. These results indicate that domain-specific fine-tuning internalizes not only the report structure but also the professional reasoning conventions of vibration analysis practice.

# 5.Conclusions

This work addressed the integration of multi-level information — from physical models through data-driven diagnostics to natural language reasoning — in intelligent bearing fault diagnosis, and proposed a framework that extends the output of a diagnoser from a label–confidence pair to a structured, per-sample evidence record. The Diagnostic Evidence Network (DENet) augments an arbitrary sequence encoder with characteristic frequency regression and transient impulse localization; paired experiments across four encoder architectures showed that this evidence is obtained at no statistically significant accuracy cost, while the frequency head recovers characteristic frequencies with an MAE of about 6 Hz from 1,024-point segments.

The central empirical finding concerns the deviation between the predicted and theoretical characteristic frequency. This quantity, computable at inference time without ground-truth labels, detects misclassifications with AUROC 0.970 and 0.871 on Paderborn and JNU (vs. 0.983 and 0.969 for softmax confidence), stratifies diagnoses by error rate, and retains separation within the high-confidence subset. Deviation-based escalation rules can therefore provide a deployment-time, per-sample consistency check that complements existing confidence-based reliability mechanisms. In the report-generation experiments, prompt-level prohibitions left 10–12% of LLM-generated reports containing unsupported claims, whereas a six-minute QLoRA adaptation reduced this rate to 2% with no fabricated quantities observed, which favors lightweight fine-tuning over prompting for the 7B model tested.

Several limitations delineate the scope of these conclusions.First, the frequency-deviation check covers only fault-class predictions. The 182 missed detections observed on Paderborn fall outside its reach, and Fig. 4(b) suggests that testing Normal predictions against the shaft-frequency anchor is unlikely to recover them, because the frequency head on these segments also reverts to that anchor. Screening such coherent errors requires evidence that does not share the encoder, which is left to future work. Second, no causal claim is made for the deviation mechanism: a large deviation may reflect an intrinsically ambiguous signal degrading both heads, which does not affect its operational value but cautions against mechanistic interpretation. Third, all three benchmarks employ artificially induced damage under laboratory conditions; validating the evidence framework on naturally degraded bearings and in-service data remains future work, as does extending the characteristic-frequency formulation to other rotating components with theoretically derivable signatures, such as gear meshing frequencies, and generalizing the multi-level integration framework to broader industrial scenarios where physical-model, data-driven, and semantic information coexist.

# Data availability statement

The CWRU dataset is available at https://engineering.case.edu/bearingdatacenter. The Paderborn dataset is available at https://groups.uni-paderborn.de/kat/BearingDataCenter/. The JNU dataset is available at https://github.com/ClarkGableWang/JNU-Bearing-Dataset. Code will be made available upon acceptance.

# Declaration of competing interest

The authors declare that they have no known competing financial interests or personal relationships that could have appeared to influence the work reported in this paper.

# Declaration of generative AI and AI-assisted technologies in the manuscript preparation process

During the preparation of this work the author used Gemini (Gemini 3.1 Pro) in order to improve the readability and language quality of the manuscript. After using this tool, the author reviewed and edited the content as needed and takes full responsibility for the content of the published article.